%% file: neurips_2026.tex
\documentclass{article}

\usepackage[preprint]{neurips_2026}

\usepackage[utf8]{inputenc}
\usepackage[T1]{fontenc}
\usepackage{hyperref}
\usepackage{url}
\usepackage{booktabs}
\usepackage{amsfonts}
\usepackage{amsmath}
\usepackage{amssymb}
\usepackage{nicefrac}
\usepackage{microtype}
\usepackage{xcolor}

\usepackage{graphicx}
\usepackage{subcaption}
\usepackage{multirow}
\usepackage{algorithm}
\usepackage{algpseudocode}
\usepackage{listings}
\usepackage{xspace}
\usepackage{tikz}
\usetikzlibrary{positioning,arrows.meta,calc,shapes.geometric,decorations.pathmorphing,fit,backgrounds}
\usepackage[most]{tcolorbox}
\tcbuselibrary{listings,breakable,skins}
\usepackage[capitalize,noabbrev]{cleveref}
\hypersetup{
  pdftitle={SePO: Self-Evolving Prompt Agent for System Prompt Optimization},
  pdfauthor={Wangcheng Tao, Han Wu, Weng-Fai Wong},
  pdfsubject={System prompt optimization},
}

\graphicspath{{figures/}}
\newcommand{\method}{SePO\xspace}
\newcommand{\methodfull}{Self-Evolving Prompt Optimization\xspace}
\newcommand{\promptagent}{prompt agent\xspace}
\newcommand{\Promptagent}{Prompt agent\xspace}

\newcommand{\ie}{i.e.,\xspace}

\definecolor{promptboxbg}{HTML}{FFF4E8}
\definecolor{promptboxframe}{HTML}{C2570B}
\newtcblisting{promptbox}[2]{%
  enhanced,
  breakable,
  colback=promptboxbg,
  colframe=promptboxframe,
  colbacktitle=promptboxframe,
  coltitle=white,
  fonttitle=\bfseries,
  title={#1},
  label={#2},
  arc=1mm,
  boxrule=0.6pt,
  left=4mm,right=4mm,top=3mm,bottom=3mm,
  listing only,
  listing options={
    basicstyle=\ttfamily\small\linespread{1.08}\selectfont,
    breaklines=true,
    breakindent=2em,
    breakautoindent=false,
    breakatwhitespace=true,
    prebreak=,
    postbreak=,
    columns=fullflexible,
    showstringspaces=false,
    keepspaces=true,
    frame=none,
    aboveskip=0pt,
    belowskip=0pt,
    upquote=true,
    extendedchars=true,
    inputencoding=utf8,
    literate=%
      {—}{{\textemdash{}}}1
      {–}{{\textendash{}}}1
      {“}{{\textquotedblleft{}}}1
      {”}{{\textquotedblright{}}}1
      {‘}{{\textquoteleft{}}}1
      {’}{{\textquoteright{}}}1
      {…}{{\ldots{}}}1,
  },
}
\crefname{tcblisting}{Listing}{Listings}
\Crefname{tcblisting}{Listing}{Listings}

\title{SePO: Self-Evolving Prompt Agent\\
for System Prompt Optimization}

\author{%
  Wangcheng Tao \\
  National University of Singapore \\
  \texttt{taowangcheng@u.nus.edu} \\
  \And
  Han Wu\thanks{Corresponding author.} \\
  City University of Hong Kong \\
  \texttt{hanwu.cs@my.cityu.edu.hk} \\
  \And
  Weng-Fai Wong \\
  National University of Singapore \\
  \texttt{wongwf@nus.edu.sg} \\
}

\begin{document}

\maketitle
\footnotetext[1]{\url{https://github.com/taowangcheng/SePO}}

\input{sections/00_abstract}

\input{sections/01_introduction}

\input{sections/02_related_work}

\input{sections/03_methodology}

\input{sections/04_experiments}

\input{sections/05_conclusion}

\bibliographystyle{plainnat}
\bibliography{references}

\newpage
\appendix
\section*{\centering\Large Appendix}
\input{sections/appendix}

\end{document}

%% file: sections/00_abstract.tex
\begin{abstract}
System prompt optimization improves agent behavior without modifying the underlying model, yielding human-readable, model-agnostic instructions.
Existing methods build a prompt agent that refines task agents' system prompts, yet leave the prompt agent's own system prompt hand-engineered and fixed.
We propose \methodfull (\method{}), which treats the prompt agent's own system prompt as an optimization target alongside task agents' system prompts.
\method{} adopts a self-referential design. A single prompt agent improves both task agents' system prompts and its own under an open-ended evolutionary search that maintains an archive of candidate prompts as stepping stones.
Training proceeds in two stages: pre-training evolves the prompt agent on a multi-task pool, and fine-tuning then applies it to a target task.
Across five benchmarks spanning math (AIME'25), abstract reasoning (ARC-AGI-1), graduate-level science (GPQA), code generation (MBPP), and logic puzzles (Sudoku), \method{} consistently outperforms Manual-CoT, TextGrad, and MetaSPO, improving the average accuracy by $4.49$ points compared to Manual-CoT.
The prompt optimization skill from pre-training also generalizes to tasks beyond the pre-training mixture, rather than memorizing per-task prompts.
\end{abstract}

%% file: sections/01_introduction.tex
\section{Introduction}
\label{sec:intro}

Agents are now widely deployed to perform specific tasks across reasoning~\citep{react2023}, coding~\citep{sweagent2024}, and decision-making~\citep{voyager2023}.
An agent's performance can be improved by retraining its model weights~\citep{instructgpt2022}, augmenting its memory~\citep{memgpt2023}, designing its workflow~\citep{adas2024, dspy2023}, or optimizing its system prompt~\citep{ape2023, opro2024, textgrad2025, metaspo2025}.
We focus on system prompt optimization, which improves agent behavior without modifying the underlying model and produces human-readable, model-agnostic instructions.

Methods for system prompt optimization span several lines of work.
Early work casts prompt search as a black-box optimization problem driven by evaluation feedback~\citep{ape2023, opro2024}.
A subsequent line of work runs evolutionary search over a population of candidate prompts~\citep{promptbreeder2023, evoprompt2024}.
More recent methods backpropagate natural language critiques through textual-gradient frameworks~\citep{textgrad2025} and meta-learn a shared cross-task prompt~\citep{metaspo2025}.
Across these methods, a \emph{prompt agent} reads evaluation feedback and proposes refined prompts for the task agent.
The prompt agent is itself hand-engineered and does not improve as more tasks are seen.
Prompt optimization is therefore bounded by what a human can hand-engineer, and does not benefit from accumulated experience.

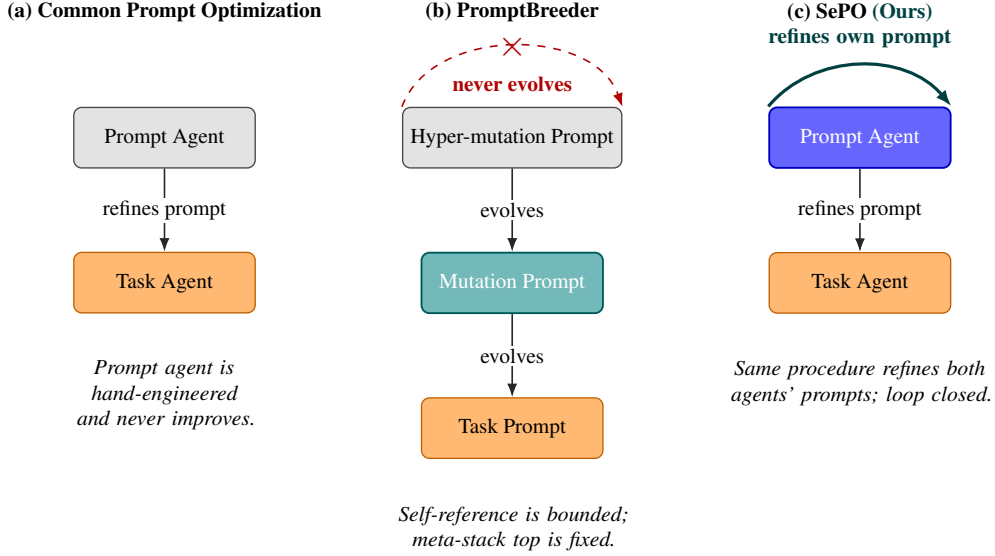
\begin{figure}[t]
  \centering
  \input{figures/fig1_loop}
  \caption{\textbf{Self-Referential Design in System Prompt Optimization.} (a) Common prompt optimization methods leave the prompt agent hand-engineered, so the optimization loop never includes the prompt agent itself. (b) PromptBreeder introduces a meta-stack but its top stays fixed, leaving the loop bounded but never closed. (c) Under \method{}'s self-referential design, the same procedure refines both task agents' system prompts and the prompt agent's own, closing the loop.}
  \label{fig:loop}
\end{figure}

The root issue is that only the task agent's prompt is treated as an optimization target, while the prompt agent itself stays fixed.
We close this gap with a self-referential design.
The prompt agent treats itself as a special task agent, so the same procedure that refines any task agent's prompt also refines its own.
\Cref{fig:loop} contrasts \method{}'s self-referential design with prior prompt optimization methods.
The procedure runs as an open-ended evolution over a population of candidate prompts, inspired by~\citet{dgm2025}.
An archive lets earlier prompts serve as stepping stones for later improvements.
We call this framework \emph{\methodfull} (\method).
The same procedure now covers both layers, removing the need to hand-design a separate optimizer for the prompt agent.
Mimicking the standard pre-training and fine-tuning paradigm, we organize the procedure into two stages.
The first stage, namely the ``pretraining'' of the prompt agent, runs the self-referential loop on a pool of tasks, evolving a strong prompt agent with the general capacity across various scenarios. The second stage, termed ``fine-tuning'' on specific tasks, leverages the prompt agent to improve the prompt of the targeted task agent.
This split amortizes the cost of self-evolving the prompt agent across many fine-tuning tasks.
Multi-task pre-training also draws on the standard principle that diverse training data improves both robustness and generalization.
Together, the self-referential design and the two-stage training pipeline turn prompt optimization from a fixed tool into a learnable skill that accumulates across tasks.

We evaluate \method{} on five benchmarks spanning math (AIME'25), abstract reasoning (ARC-AGI-1), graduate-level science (GPQA), code generation (MBPP), and logic puzzles (Sudoku).
Against three prompt optimization baselines (Manual-CoT, TextGrad, MetaSPO), \method{} achieves the best accuracy on every task, improving the average accuracy by $4.49$ points compared to Manual-CoT.
Splitting the training into the pre-training and fine-tuning stages also gives a clean separation of concerns.
Pre-training runs once on a multi-task pool, and the resulting prompt agent is then reused across various task agents during fine-tuning.
The prompt optimization skill from pre-training also extends to tasks beyond the pre-training mixture, rather than being memorized per task.

%% file: figures/fig1_loop.tex
\begin{tikzpicture}[
    font=\footnotesize,
    box/.style={draw, rounded corners=3pt, align=center, minimum height=8mm, minimum width=24mm, inner sep=3pt, line width=0.5pt},
    learnbox/.style={box, fill=blue!60, text=white, line width=0.7pt, draw=blue!75!black},
    fixedbox/.style={box, fill=gray!22, text=black, line width=0.5pt, draw=gray!55!black},
    taskbox/.style={box, fill=orange!55, text=black, line width=0.5pt, draw=orange!75!black},
    mutbox/.style={box, fill=teal!55, text=white, line width=0.7pt, draw=teal!70!black},
    arrow/.style={-{Latex[length=1.8mm]}, line width=0.6pt, draw=gray!30!black},
    closeloop/.style={-{Latex[length=2.2mm]}, line width=1.2pt, draw=teal!50!black},
    failloop/.style={-{Latex[length=1.8mm]}, line width=0.6pt, dashed, draw=red!70!black},
    panellabel/.style={font=\bfseries\footnotesize, align=center},
    arrlabel/.style={font=\footnotesize, align=center, fill=white, inner sep=1pt},
    closelabel/.style={font=\footnotesize\bfseries, align=center, color=teal!50!black, text width=28mm},
    faillabel/.style={font=\footnotesize\bfseries, align=center, color=red!70!black},
    captionrow/.style={font=\footnotesize\itshape, align=center, text width=36mm}
]

\def\panelsep{46mm}
\def\boxgap{11mm}

\node[panellabel] (a-title) at (0, 0) {(a) Common Prompt Optimization};
\node[fixedbox, below=10mm of a-title] (a-pa) {Prompt Agent};
\node[taskbox, below=\boxgap of a-pa] (a-ta) {Task Agent};
\draw[arrow] (a-pa) -- (a-ta) node[arrlabel, midway] {refines prompt};
\node[captionrow, below=5mm of a-ta] {Prompt agent is hand-engineered\\and never improves.};

\node[panellabel] (b-title) at (\panelsep, 0) {(b) PromptBreeder};
\node[fixedbox, below=10mm of b-title] (b-hm) {Hyper-mutation Prompt};
\node[mutbox, below=\boxgap of b-hm] (b-mp) {Mutation Prompt};
\node[taskbox, below=\boxgap of b-mp] (b-tp) {Task Prompt};
\draw[arrow] (b-hm) -- (b-mp) node[arrlabel, midway] {evolves};
\draw[arrow] (b-mp) -- (b-tp) node[arrlabel, midway] {evolves};
\draw[failloop] (b-hm.north west) to[bend left=70] (b-hm.north east);
\node[font=\bfseries, color=red!70!black, scale=1.4] at ($(b-hm.north)+(0,8mm)$) {$\times$};
\node[faillabel] at ($(b-hm.north)+(0,3mm)$) {never evolves};
\node[captionrow, below=5mm of b-tp] {Self-reference is bounded;\\meta-stack top is fixed.};

\node[panellabel] (c-title) at (2*\panelsep, 0) {(c) \method{} \textcolor{teal!50!black}{(Ours)}};
\node[learnbox, below=10mm of c-title] (c-pa) {Prompt Agent};
\node[taskbox, below=\boxgap of c-pa] (c-ta) {Task Agent};
\draw[arrow] (c-pa) -- (c-ta) node[arrlabel, midway] {refines prompt};
\draw[closeloop] (c-pa.north west) to[bend left=50] node[closelabel, above=3pt] {refines own prompt} (c-pa.north east);
\node[captionrow, below=5mm of c-ta] {Same procedure refines both\\agents' prompts; loop closed.};

\end{tikzpicture}

%% file: sections/02_related_work.tex
\section{Related Work}
\label{sec:related}

\paragraph{Prompt Optimization}
Prompt optimization has received considerable attention since chain-of-thought prompting~\citep{cot2022} demonstrated that simple structural cues can substantially improve agent reasoning.
These cues were initially hand-crafted, a labor-intensive process that motivated subsequent work on automated prompt optimization.
Early black-box methods treat prompt search as an optimization problem driven by an agent that reads evaluation feedback and proposes refined prompts~\citep{ape2023, opro2024}.
This approach struggles when good prompts are sparse, leading to evolutionary methods that maintain a population of candidates and apply mutation and selection~\citep{promptbreeder2023, evoprompt2024}.
A separate line of work approaches prompt optimization through established machine-learning paradigms, moving beyond heuristic search.
Textual-gradient frameworks~\citep{textgrad2025} propagate natural language critiques through agent compute graphs, providing component-level feedback rather than population-level fitness.
Meta-learning~\citep{metaspo2025} instead produces a shared cross-task prompt, generalizing optimization across tasks.
Among these, PromptBreeder is the closest precedent to \method{}.
It co-evolves task prompts alongside the mutation prompts producing them, an early form of self-referential prompt evolution.
The self-reference is nevertheless bounded. A hand-written hyper-mutation prompt evolves the mutation prompt and is itself never evolved.
The meta-stack therefore has a fixed hand-engineered top, and the loop is never closed (\Cref{fig:loop}b).
Each evolutionary run is also task-specific, since task and mutation prompts are coupled into one unit and re-initialized per task.
MetaSPO aligns most directly with our problem framing, formulating prompt optimization as cross-task meta-learning.
The meta-optimizer itself, however, remains hand-written and outside the meta-learning loop.
Across the methods above, the prompt agent driving the search is itself hand-engineered and does not improve as more tasks are seen.
To address this, \method{} treats the prompt agent as a special task agent, so the same procedure refines both prompts.
An archive lets earlier prompts serve as stepping stones for later improvements.
A pre-training stage runs the search on a diverse multi-task pool, and fine-tuning then reuses the resulting prompt agent on various tasks.

\paragraph{Self-Evolving Agents}
Self-evolving agents can be classified by what they modify and at what stage~\citep{selfevolvingsurvey2026}.
The earliest self-improvement methods modify only the agent's outputs.
Self-Refine~\citep{selfrefine2023} and Reflexion~\citep{reflexion2023} have agents produce structured natural-language critiques of their own outputs and incorporate them into subsequent attempts.
Moving beyond per-output revision, Voyager~\citep{voyager2023} accumulates a library of skills usable across episodes in an open-ended Minecraft environment.
Most recently, work has moved to the agent's code and architecture rather than its outputs.
Archive-based evolutionary search over coding agents~\citep{dgm2025} maintains a population of agent variants evaluated on a fixed benchmark.
This produces open-ended self-improvement, building on the original G\"odel Machine proposal of~\citet{godel2003}.
ADAS~\citep{adas2024} similarly evolves agent system code, with a fixed meta-agent generating candidate designs.
Within this lineage, \method{} operates only on the prompt agent's natural language system prompt, leaving code, weights, and tools untouched.
The system prompt is more interpretable and model-agnostic than agent outputs, accumulated skills, or agent architectures.

\paragraph{Evolutionary Search over Non-Agent Artifacts}
A parallel line of work runs evolutionary search over artifacts that are not themselves agents.
FunSearch~\citep{funsearch2024} evolves programs that produce mathematical objects, achieving new combinatorial bounds.
Eureka~\citep{eureka2024} extends the template to reinforcement learning, refining reward functions through agent-proposed code.
AlphaEvolve~\citep{alphaevolve2025} scales the same idea to algorithms for scientific and engineering problems.
Across these systems, the agent driving the search is a fixed external operator, separate from the artifacts being evolved.
\method{}, by contrast, places the prompt agent itself inside the population it searches over, so the operator is itself a target of optimization.

%% file: sections/03_methodology.tex
\section{Methodology}
\label{sec:method}

We first formalize the notions of tasks and agents, then state the standard problem of system prompt optimization for a task agent.
We then map the prompt agent's own system prompt to the same problem and propose \method{}, which optimizes both prompts within a single procedure across two training stages.

\begin{figure}[t]
  \centering
  \includegraphics[width=\linewidth]{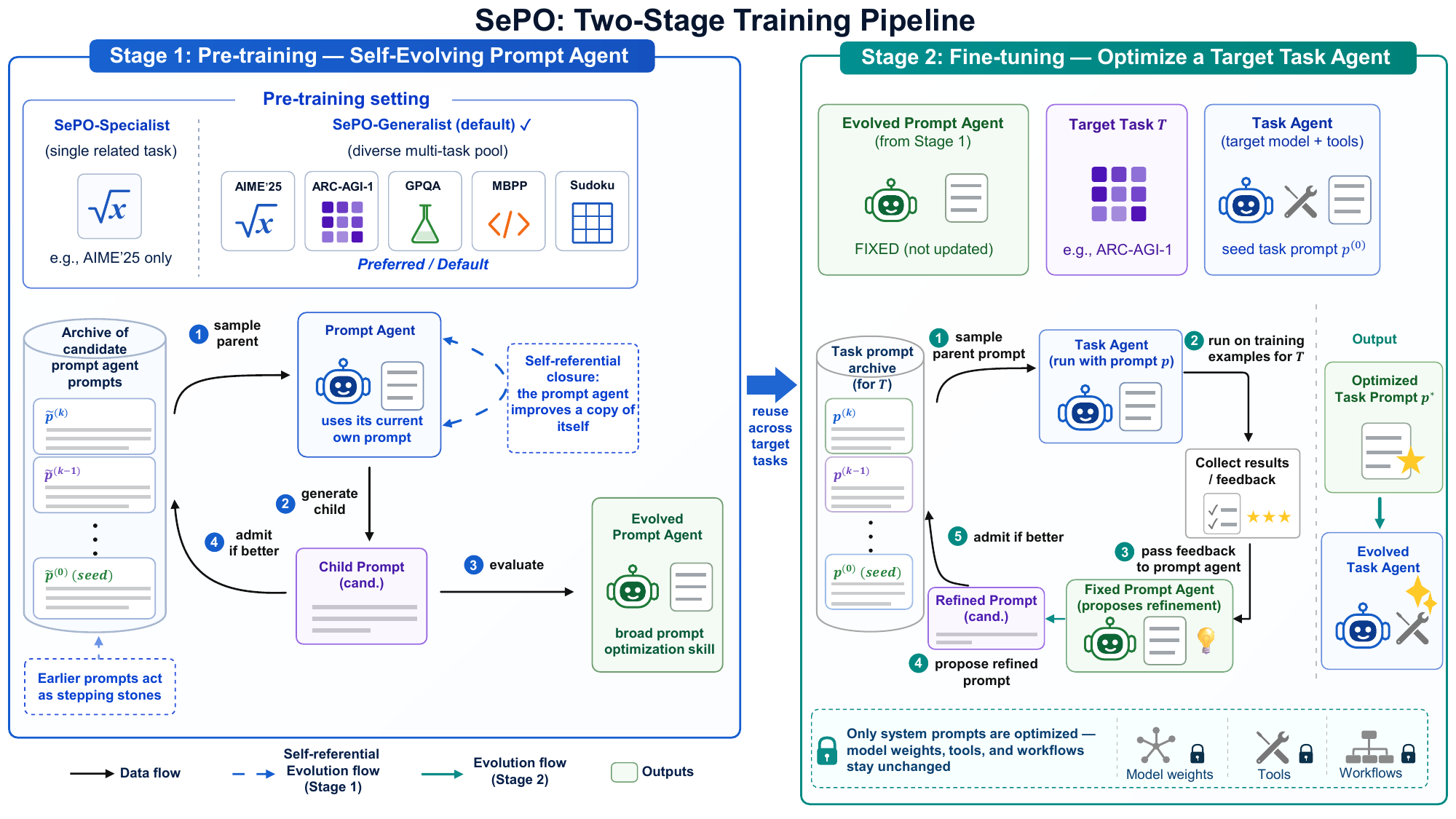}
  \caption{\textbf{Overview of \method{}'s Two-Stage Training Pipeline.} \textbf{Pre-training} (left) evolves the prompt agent's own system prompt $\tilde{p}$ through open-ended evolutionary search, maintaining an archive of candidate prompts as stepping stones. The pre-training task pool is either a single task (\method{}-Specialist) or a multi-task mixture (\method{}-Generalist; see \Cref{sec:method:generalist}). \textbf{Fine-tuning} (right) reuses the resulting $\tilde{p}^{\star}$ to optimize a task agent's system prompt $p$ on a target task, again through open-ended evolutionary search.}
  \label{fig:pipeline}
\end{figure}

\subsection{Preliminary}
\label{sec:method:prelim}

\paragraph{Tasks and Task Agents}
A \emph{task} $T = (\mathcal{D}, S)$ is a dataset $\mathcal{D}$ of input--target pairs $(x, y)$ together with a deterministic scoring function $S(x, y, \hat{y})$.
A \emph{task agent} takes a task input $x$ and returns a candidate response $\hat{y}$.
We write it as a tuple $A = A_{(p, M, W)}$ comprising a system prompt $p$, an underlying language model $M$, and a workflow $W$ that wraps $x$ into a user prompt, queries $M$, and parses the response.
Applying the agent is shorthand for $A(x) = W_{M}(x \mid p)$.
The accuracy of $A$ on a task $T$ is $\mathrm{acc}(A; T) = \mathbb{E}_{(x, y) \sim \mathcal{D}}[S(x, y, A(x))]$.

\paragraph{Prompt Agents and Standard System Prompt Optimization}
The standard problem of system prompt optimization for a task agent $A$ on task $T$ is to find
\begin{equation}
\label{eq:standard_spo}
p^{\star} = \arg\max_{p}\; \mathbb{E}_{(x, y) \sim \mathcal{D}}\!\left[S\!\left(x,\; y,\; A_{(p, M, W)}(x)\right)\right].
\end{equation}
Prior work~\citep{ape2023, opro2024, textgrad2025, metaspo2025} solves this by introducing a second agent, the \emph{prompt agent}, $\tilde{A}$, that reads the evaluation feedback and proposes refined task agent prompts.
The prompt agent has the same form as a task agent, with its own system prompt, $\tilde{p}$, model, and workflow.
Its input is a tuple $\tilde{x} = (T, A, E)$ comprising a task $T$, the task agent $A$ being optimized, and a batch of evaluation results $E$ from running $A$ on $T$.
After invocation, the prompt agent produces a refined prompt $p' = \tilde{A}(\tilde{x})$ expected to score higher than $p$.
When $T$ and the non-prompt components of $A$ are fixed within a run, $\tilde{A}$ depends only on $p$ and $E$, and we equivalently write $p' = \tilde{A}(p, E)$.
Iterating produces a sequence $p^{(0)}, p^{(1)}, \ldots$, and the highest-scoring prompt is taken as $p^{\star}$.

\paragraph{An Asymmetric Optimization}
The prompt agent is itself a task agent. Its task is to improve task agent prompts.
Yet in existing prompt optimization methods, the prompt agent's system prompt, i.e., $\tilde{p}$, is hand-engineered and fixed, leaving only the task agent's system prompt $p$ as a learnable component.
This limits the optimization of $\tilde{p}$ to whatever the human author can produce.

\subsection{Self-Referential System Prompt Optimization}
\label{sec:method:selfref}

We treat $\tilde{p}$ as an optimization variable in the same way as $p$.
To do so, we define a \emph{prompt task} $\tilde{T} = (\tilde{\mathcal{D}}, \tilde{S})$.
$\tilde{\mathcal{D}}$ collects prompt agent inputs $\tilde{x} = (T, A, E)$ as defined in \Cref{sec:method:prelim}, with $T$ being a pool of tasks.
The scoring function $\tilde{S}$ measures whether the refined prompt $p'$ improves task agent accuracy over $p$ on $T$.
With the prompt task in place, the optimization of the prompt agent's system prompt becomes
\begin{equation}
\label{eq:selfref}
\tilde{p}^{\star} = \arg\max_{\tilde{p}}\; \mathbb{E}_{(T, A, E) \sim \tilde{\mathcal{D}}}\!\left[\tilde{S}\!\left(T,\; A,\; E,\; \tilde{A}_{(\tilde{p}, \tilde{M}, \tilde{W})}(T, A, E)\right)\right].
\end{equation}
\Cref{eq:selfref} is essentially \Cref{eq:standard_spo} with a different set of parameters to optimize an agent's system prompt so as to maximize its accuracy on a task.
The same procedure therefore applies to both: improving the task agent's $p$ on $T$, and the prompt agent's $\tilde{p}$ on $\tilde{T}$.
We call this the \emph{self-referential closure}: the prompt agent treats itself as a special task agent whose task is to improve task agent prompts.

\subsection{\method{}: \methodfull{}}
\label{sec:method:sepo}

\emph{\methodfull{}} (\method) instantiates the optimizations of \Cref{eq:standard_spo} and \Cref{eq:selfref} by applying one search algorithm at two training stages.
The search is an open-ended evolution over a population of candidate prompts, inspired by~\citet{dgm2025}.
It maintains an archive of candidates and admits children who improve on their parents.
The archive lets earlier prompts serve as stepping stones for later improvements.
For implementation details, please refer to Appendix~\ref{app:oees}.

\paragraph{Two-Stage Training Pipeline}
\Cref{fig:pipeline} provides an overview of the pipeline.
We train \method{} in two stages, mirroring the standard pre-training and fine-tuning paradigm. The two stages share \Cref{alg:sepo}, differing only in which agent's system prompt is the optimization target.
\textbf{Pre-training} first equips the prompt agent with broad prompt optimization skill across many tasks. It evolves the prompt agent's own system prompt from seed $\tilde{p}^{(0)}$ to $\tilde{p}^{\star}$ on the prompt task $\tilde{T}$. The self-referential closure activates here: the parent $\tilde{p}$ at each generate step is itself the prompt agent's system prompt, so the agent improves a copy of itself.
\textbf{Fine-tuning} then applies this prompt agent to optimize a task agent's prompt for a target task. It evolves a task agent's system prompt from $p^{(0)}$ to $p^{\star}$ on a single task $T$. During fine-tuning, the prompt agent uses $\tilde{p}^{\star}$ as its system prompt throughout.
Together, this two-stage training pipeline amortizes a single pre-training run across many fine-tuning tasks. The prompt agent accumulates broad prompt optimization skill from the multi-task pool, supporting cross-task generalization.

\begin{algorithm}[t]
\caption{\methodfull{}}
\label{alg:sepo}
\begin{algorithmic}[1]
\Require Stage $\in \{\text{pre-training}, \text{fine-tuning}\}$, seed prompt $p^{(0)}$, generations $G$, children per generation $K$. During fine-tuning, the prompt agent uses a fixed $\tilde{p}^{\star}$ (output of pre-training).
\Ensure Best evolved prompt $p^{\star}$.
\State $\mathcal{A} \gets \{p^{(0)}\}$
\For{$t = 1, \dots, G$}
    \For{$k = 1, \dots, K$ \textbf{in parallel}}
        \State Sample parent $p \in \mathcal{A}$
        \State Generate child $p'$ via the prompt agent \Comment{prompt agent's system prompt: $p$ during pre-training, $\tilde{p}^{\star}$ during fine-tuning}
        \State Compute score $s_{p'}$ and admit $p'$ to $\mathcal{A}$ if it passes the admission criterion
    \EndFor
\EndFor
\State \Return $p^{\star} \gets \arg\max_{p \in \mathcal{A}} s_p$
\end{algorithmic}
\end{algorithm}

\paragraph{\method{}-Specialist vs.\ \method{}-Generalist}\label{sec:method:generalist}
The two configurations of \method{} differ only in which tasks the prompt agent sees during pre-training.
\textbf{\method{}-Specialist} uses a single task, \ie{} the same task fine-tuning later targets.
\textbf{\method{}-Generalist} uses a mixture of multiple tasks, drawing on the standard pre-training paradigm in which diverse training data improves both robustness and generalization.
A single pre-training run can also be reused across many fine-tuning tasks, whereas \method{}-Specialist must run pre-training separately for each.
\method{}-Generalist is our default configuration, whereas \method{}-Specialist is the single-task variant.

\paragraph{Task Selection for \method{}-Generalist}
\method{}-Generalist needs to choose which tasks to include in its pre-training mixture.
Multi-task prompt optimization addresses this problem, with methods such as dynamic task grouping~\citep{dtvg2025}, meta-learned task selection~\citep{metaspo2025}, and high-variance subset selection~\citep{p1prompt2026}.
Since this is not the focus of \method{}, we use a greedy heuristic (Appendix~\ref{app:generalist-algo}).
More advanced task selection algorithms could replace our heuristic without affecting the rest of \method{}.

%% file: sections/04_experiments.tex
\section{Experiments}
\label{sec:experiments}

We evaluate \method{} on five tasks against three prompt optimization baselines.
\Cref{sec:setup} introduces the tasks, baselines, our two \method{} configurations, and implementation details.
\Cref{sec:results} reports main results, validates the multi-task selection heuristic, ablates \method{}'s components, probes cross-task generalization and robustness across model pairs, compares training cost across methods, and analyzes the evolved prompts qualitatively.

\subsection{Setup}
\label{sec:setup}

\paragraph{Tasks}
\label{sec:tasks}
Our evaluation suite spans five tasks with distinct skill profiles: mathematics, abstract visual reasoning, graduate-level science, code synthesis, and combinatorial puzzles.
Each task is defined by a dataset $\mathcal{D}$ and a scoring function $S$ following \Cref{sec:method:prelim}.
\textbf{AIME'25}~\citep{aime2025hf} poses high-school olympiad mathematics problems with integer answers in $[0, 999]$.
We extract the boxed integer from the task agent's response and check it against the gold integer.
\textbf{ARC-AGI-1}~\citep{arcagi2019} tests abstract visual reasoning over grid-to-grid transformation puzzles.
We compare each predicted output grid to the reference.
\textbf{GPQA}~\citep{gpqa2023} covers graduate-level multiple-choice science questions across physics, chemistry, and biology.
We match the predicted answer letter against the gold key.
\textbf{MBPP}~\citep{mbpp2021} requires Python program synthesis from a docstring against hidden unit tests.
We execute the predicted program against those tests and score by functional correctness.
\textbf{Sudoku}~\citep{d1diff2025} presents $4{\times}4$ Sudoku puzzles.
We score the predicted grid using a Sudoku validator that checks the row, column, and subgrid constraints.
We report pass@3 accuracy for ARC-AGI-1 following the official protocol and pass@1 accuracy for the other four tasks.
We partition each $\mathcal{D}$ into a train split for prompt optimization and a test split for the final evaluation in \Cref{sec:results}.
For a more accurate estimate, we evaluate each test problem multiple times and report the average accuracy.
More task details are provided in Appendix~\ref{app:task-construction}.

\paragraph{Baselines}
\label{sec:baselines}
We compare \method{} against three baselines that span the spectrum from no automatic optimization to a recent meta-learning method.
\textbf{Manual-CoT} is a no-optimization baseline that uses a hand-crafted system prompt with CoT-style task-handling guidance (``think step by step'').
This baseline isolates the contribution of any automatic prompt optimization.
\textbf{TextGrad}~\citep{textgrad2025} is a text-gradient framework that backpropagates natural language critique through an autograd graph over agent calls.
We adapt the official implementation to use our model registry and per-task evaluation harness.
The prompt optimization component in TextGrad is hand-engineered and fixed throughout optimization, in contrast to the self-evolving prompt agent in \method{}.
\textbf{MetaSPO}~\citep{metaspo2025} meta-learns a cross-task global system prompt by alternating outer-loop system-prompt optimization with inner-loop user-prompt refinement.
We use the authors' published global prompt and append per-task descriptions and answer-format constraints before evaluation.
We compare against MetaSPO because it is the strongest existing cross-task system-prompt baseline and targets the same cross-task generalization as \method{}-Generalist.
The meta-optimizer is also hand-written and not part of the meta-learning loop.

\paragraph{Our Method}
\label{sec:ourmethod}
\method{} runs an open-ended evolutionary search at two training stages, optimizing both the prompt agent's system prompt and a task agent's system prompt.
Both pre-training and fine-tuning use only the train split of each task.
We evaluate two configurations of \method{} introduced in \Cref{sec:method:generalist}.
\textbf{\method{}-Specialist} runs pre-training on the same single task that fine-tuning later targets.
\textbf{\method{}-Generalist} runs pre-training on a multi-task mixture selected by a greedy heuristic.
The resulting prompt agent is then reused across all fine-tuning tasks.
\method{}-Generalist is our default, while \method{}-Specialist isolates the contribution of multi-task pre-training.

\paragraph{Implementation Details}
\label{sec:impl}
Across all methods, we use DeepSeek-V3.2~\citep{deepseekv32_2025} and Gemini 3.1 Pro Preview~\citep{gemini3pro_2026} as the underlying models for the task agent and prompt agent, respectively.
We use temperature $0$ for the task agent to keep its responses deterministic and the evaluation accurate.
For the prompt agent we use temperature $1$, encouraging diverse candidate prompts during prompt optimization.
TextGrad and \method{} share the same optimization budget of $10$ iterations with $16$ examples per iteration.
At both pre-training and fine-tuning, \method{} runs $G{=}5$ generations of $K{=}2$ children.
Each child is evaluated against a batch of failed and successful task examples at a $\sim$1:1 ratio.
All reported results are averaged over $5$ independent runs with different random seeds.
Per-method implementation details, including each baseline's adaptation to our harness, are in Appendix~\ref{app:impl-details}.

\subsection{Results and Analyses}
\label{sec:results}

\paragraph{Main Results}
\label{sec:results:main}

We first report the performance of \method{} against three prompt optimization baselines on the five evaluation tasks.
\Cref{tab:main} shows that \method{}-Generalist achieves the best accuracy on every task, improving the average accuracy from $71.89$ (Manual-CoT) to $76.38$.
\method{}-Specialist also improves over Manual-CoT on every task but trails \method{}-Generalist by $2.29$ points in average accuracy.
TextGrad and MetaSPO each fall below Manual-CoT on at least three tasks, with average accuracies of $70.39$ and $71.32$, respectively.
TextGrad's prompt optimization component is hand-engineered and fixed throughout optimization, so it cannot adapt as the search proceeds.
MetaSPO inherits the same fixed-optimizer limitation and additionally meta-learns a single cross-task global system prompt, an artifact that cannot match per-task reasoning needs.
\method{} instead evolves the prompt agent itself across two training stages.
Cross-task generalization therefore comes from the prompt optimization skill, not from a hand-written optimizer or a memorized global prompt.
These results demonstrate that self-improvement of the prompt agent raises the ceiling above what hand-written prompt optimization methods alone can reach.

\begin{table}[t]
  \centering
  \small
  \caption{\textbf{Main Results across Five Evaluation Tasks.} Per-task test accuracy ($\uparrow$) of \method{}-Specialist and \method{}-Generalist against three prompt optimization baselines. DeepSeek-V3.2 was used for the task agent, and Gemini 3.1 Pro Preview for the prompt agent. The `Avg.' column gives the average across the five tasks. Best per column in \textbf{bold}.}
  \label{tab:main}
  \begin{tabular}{lcccccc}
    \toprule
    Method & AIME'25 & ARC-AGI-1 & GPQA & MBPP & Sudoku & Avg. \\
    \midrule
    Manual-CoT & 57.55 & 37.30 & 76.46 & 91.20 & 96.95 & 71.89 \\
    TextGrad   & 55.99 & 34.75 & 74.44 & 90.15 & 96.60 & 70.39 \\
    MetaSPO    & 57.71 & 37.27 & 75.51 & 89.30 & 96.80 & 71.32 \\
    \midrule
    \method{}-Specialist & 60.94 & 37.46 & 76.72 & 95.55 & 99.80 & 74.09 \\
    \method{}-Generalist & \textbf{64.22} & \textbf{43.39} & \textbf{78.18} & \textbf{96.20} & \textbf{99.90} & \textbf{76.38} \\
    \bottomrule
  \end{tabular}
\end{table}

\paragraph{Task Selection for \method{}-Generalist}
\label{sec:results:generalist}

\begin{figure}[t]
  \centering
  \begin{minipage}[t]{0.48\textwidth}
    \centering
    \includegraphics[width=\linewidth]{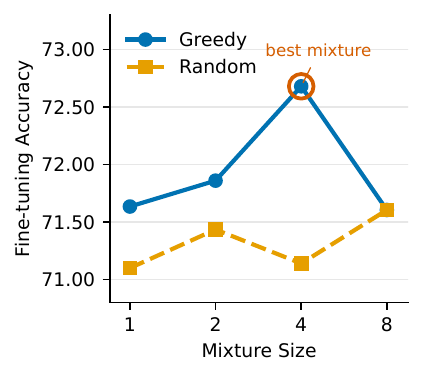}
    \caption{\textbf{Greedy vs.\ Random Task Selection.} Average fine-tuning accuracy on the AIME'25 and ARC-AGI-1 \emph{training splits}, for pre-training task mixtures of sizes $\{1, 2, 4, 8\}$ chosen by a \emph{Greedy} or \emph{Random} selector. The highlighted point is the best mixture, the size-4 greedy \texttt{STEM+ARC-AGI-1+LIMO+MBPP}.}
    \label{fig:generalist}
  \end{minipage}\hfill
  \begin{minipage}[t]{0.48\textwidth}
    \centering
    \includegraphics[width=\linewidth]{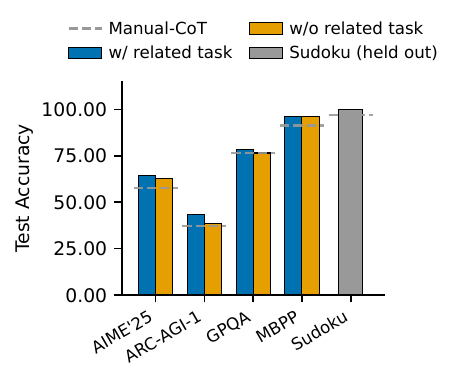}
    \caption{\textbf{Generalization with and without Related Pre-Training Tasks.} Per-task test accuracy of \method{}-Generalist under two pre-training settings: \emph{w/ related task} when the pre-training mixture contains a task related to the target, and \emph{w/o related task} when it does not. Sudoku is held out of every pre-training mixture (gray bar). \emph{Manual-CoT} shown as a dashed reference line.}
    \label{fig:generalization}
  \end{minipage}
\end{figure}

Multi-task prompt optimization~\citep{dtvg2025, metaspo2025, p1prompt2026} demonstrates that the composition of the task mixture shapes the prompt agent's optimization skill during fine-tuning.
Among the five evaluation tasks, AIME'25 and ARC-AGI-1 are the hardest and most specialized, and therefore demand sharper prompt optimization than the others.
We score each candidate mixture by the average fine-tuning accuracy across the AIME'25 and ARC-AGI-1 training splits.
\Cref{fig:generalist} compares our greedy task selector (\Cref{sec:method:generalist}) against a random selector at four mixture sizes.
The greedy selector outperforms the random selector at every size below 8, with the largest gap at size 4 ($72.68$ versus $71.14$).
At size 8, both selectors necessarily return the full pool of eight tasks.
Size 4 also matches intuition. Smaller pools omit complementary tasks, while size 8 dilutes the scoring signal with less aligned ones.
A larger mixture also makes it harder for the prompt agent to consolidate a single optimization skill across all member tasks.
The greedy selector's consistent advantage over the random selector validates the heuristic introduced in \Cref{sec:method:generalist}.
More advanced multi-task selection algorithms~\citep{dtvg2025, p1prompt2026} could plug in here for further gains, without changing the rest of \method{}.
Unless otherwise stated, we report \method{}-Generalist results with the size-4 greedy mixture \texttt{STEM+ARC-AGI-1+LIMO+MBPP}.

\paragraph{Cross-Task Generalization}
\label{sec:results:generalization}

We test whether the prompt optimization skill from pre-training generalizes to tasks beyond the pre-training mixture.
For each evaluation task, \Cref{fig:generalization} compares two pre-training settings: with and without a related task in the pre-training mixture.
For example, AIME'25 is related to STEM and LIMO since all three involve math knowledge.
On every task, even the unrelated mixture still beats Manual-CoT.
The largest gap is on ARC-AGI-1 ($+4.95$ points), the most specialized task in our suite, where a related pre-training task helps most.
The remaining tasks show much smaller gaps of $+1.30$ on AIME'25 and $+1.66$ on GPQA, with comparable MBPP accuracy across the two settings.
On these tasks, general prompt optimization skill from pre-training already accounts for most of the gain.
Yet, more interestingly, Sudoku never appears in any pre-training mixture and \method{}-Generalist still improves it from $96.95$ (Manual-CoT) to $99.90$.
This is consistent with Sudoku's relatively low specialization. General prompt optimization skill alone suffices to raise Sudoku accuracy substantially without any related pre-training data.
Overall, these results highlight \method{}'s cross-task generalization. The pre-training stage learns a generalizable prompt optimization skill rather than memorizing per-task prompts.

\paragraph{\method{} Variants}
\label{sec:results:ablation}

Recall that \method{} combines self-improvement of the prompt agent with open-ended evolution.
To verify that each component is necessary, we consider two variants.
\textbf{\method{} w/o self-improvement} skips pre-training entirely, so the prompt agent uses its hand-written seed during fine-tuning.
\textbf{\method{} w/o open-ended evolution} replaces the archive-based search with a linear search that always picks the latest candidate.
In \Cref{tab:ablation}, both variants underperform \method{}-Generalist on average ($-1.44$ and $-3.74$ points respectively).
Each variant hits a different task hardest.
W/o self-improvement hurts ARC-AGI-1 most ($-3.63$ points), while w/o open-ended evolution drops AIME'25 most ($-6.98$ points).
Both components are therefore essential to \method{}.
Self-improvement strengthens the prompt agent itself, and open-ended evolution lets the search escape local optima that linear search cannot.

\begin{table}[t]
  \centering
  \small
  \caption{\textbf{Component Ablations of \method{}-Generalist.} Per-task test accuracy under two ablations of \method{}-Generalist: removing self-improvement of the prompt agent (\emph{w/o Self-Improvement}) and replacing archive-based open-ended evolution with linear search (\emph{w/o Open-Ended Evolution}). Best per column in \textbf{bold}.}
  \label{tab:ablation}
  \begin{tabular}{lcccccc}
    \toprule
    Method & AIME'25 & ARC-AGI-1 & GPQA & MBPP & Sudoku & Avg. \\
    \midrule
    Manual-CoT               & 57.55 & 37.30 & 76.46 & 91.20 & 96.95 & 71.89 \\
    \midrule
    \method{}-Generalist     & \textbf{64.22} & \textbf{43.39} & \textbf{78.18} & \textbf{96.20} & 99.90 & \textbf{76.38} \\
    w/o Self-Improvement     & 62.81 & 39.76 & 76.21 & 95.95 & \textbf{99.95} & 74.94 \\
    w/o Open-Ended Evolution & 57.24 & 41.58 & 73.74 & 91.10 & 99.55 & 72.64 \\
    \bottomrule
  \end{tabular}
\end{table}

\paragraph{Analysis with Varying Models}
\label{sec:results:robust}

The experiments so far used DeepSeek-V3.2 and Gemini~3.1~Pro~Preview as the underlying models for the task agent and prompt agent, respectively.
To study whether the gain extends to other model pairs, we swap the underlying models.
Specifically, we use Gemini~3.1~Flash-Lite~Preview~\citep{gemini3flashlite_2026} and Claude Opus 4.6~\citep{claudeopus46_2026} for the task agent and prompt agent, respectively.
After rerunning all five tasks, \method{}-Generalist again outperforms Manual-CoT on every task in \Cref{tab:modelswap}.
Average accuracy improves from $67.95$ to $70.08$, a gain of $+2.13$ points.
\method{} therefore generalizes to various models, not just the default pair.

\begin{table}[t]
  \centering
  \small
  \caption{\textbf{Model-Swap Robustness Across All Five Tasks}. Gemini 3.1 Flash-Lite Preview used for the task agent and Claude Opus 4.6 for the \promptagent{}. The $\Delta$ row reports \method{}-Generalist minus Manual-CoT, in accuracy points.}
  \label{tab:modelswap}
  \begin{tabular}{lcccccc}
    \toprule
    Method & AIME'25 & ARC-AGI-1 & GPQA & MBPP & Sudoku & Avg. \\
    \midrule
    Manual-CoT           & 44.53    & 30.01    & 75.66    & 90.45    & 99.10    & 67.95 \\
    \method{}-Generalist & 45.89    & 32.70    & 77.17    & 95.00    & 99.65    & 70.08 \\
    \midrule
    $\Delta$             & $+1.36$  & $+2.69$  & $+1.51$  & $+4.55$  & $+0.55$  & $+2.13$ \\
    \bottomrule
  \end{tabular}
\end{table}

\paragraph{Cost}
\label{sec:results:cost}

We compare the per-task training cost of TextGrad, \method{}-Specialist, and \method{}-Generalist in \Cref{tab:cost-textgrad,tab:cost-sepo}, with full per-stage token counts in Appendix~\ref{app:cost}.
TextGrad spends \$14.75--\$26.52 per task in a single training stage, and \method{}-Specialist spends \$5.72--\$37.63 per task across pre-training and fine-tuning combined, comparable in scale.
\method{}-Generalist instead amortizes a single \$37.14 pre-training run across all five tasks, lowering the per-task average to \$7.43 (pre-training, amortized) plus \$2.41--\$15.51 (fine-tuning).
This makes \method{}-Generalist cheaper than \method{}-Specialist on every task and comparable to TextGrad on three of five tasks, while outperforming both on accuracy (\Cref{tab:main}).
Across stages and methods, the prompt agent accounts for only a small share of token volume; it consumes summarized failure cases and emits short candidate prompts, so the task agent's evaluation passes over the training and test splits dominate the cost.
These training costs are paid once before deployment, and inference cost at query time is identical across all methods, since each runs the same task agent on the same per-query budget.

\paragraph{Qualitative Results}
\label{sec:results:qual}

We contrast the seed and best-evolved prompts for the prompt agent and the five task agents in Appendices~\ref{app:prompts} and~\ref{app:taskprompts}.
The evolved prompt agent prompt expands the seed workflow with five defensive principles that guard against chain-of-thought truncation, sacrificed rigor, and overfitting to specific test cases.
The evolved task agent prompts share a common pattern across tasks.
Each preserves the role and answer format from the seed, then adds a multi-step procedural workflow targeting failure modes specific to the task.
For example, the MBPP prompt blocks global-namespace collisions on \texttt{max}/\texttt{min}, the Sudoku prompt verifies against tokenizer-induced character drops, and the ARC-AGI-1 prompt enforces coordinate transcription rules.
This task-specific scaffolding, rather than wholesale rewriting of the seed, follows the same preserve-what-works principle that the prompt agent applies to itself.

%% file: sections/05_conclusion.tex
\section{Conclusion}
\label{sec:conclusion}

In this paper, we closed a gap in system prompt optimization. Existing methods leave the prompt agent's own system prompt hand-engineered and fixed.
To address this, we proposed \methodfull (\method), which adopts a self-referential design. The prompt agent improves its own system prompt under the same procedure it applies to task agents.
This turns the prompt agent from a hand-engineered fixture into a learnable component that strengthens with experience.
\method{}'s training is split into two stages: pre-training evolves the prompt agent on a multi-task pool, and fine-tuning then applies it to each target task.
The split amortizes pre-training cost across many applications while accumulating broad prompt optimization skill for cross-task generalization.
On five tasks across math, abstract reasoning, graduate-level science, code generation, and logic puzzles, \method{} consistently outperforms Manual-CoT, TextGrad, and MetaSPO, improving the average accuracy by $4.49$ points compared to Manual-CoT.
Beyond system prompts, the same self-referential design and two-stage training pipeline could in principle improve an agent's tools, workflows, and broader scaffolds.
We hope \method{} encourages further work on self-evolving agents that continuously broaden the scope of what they can improve about themselves.

%% file: sections/appendix.tex
\section{Open-Ended Evolutionary Search Details}
\label{app:oees}

\method{} adopts the open-ended evolutionary search of~\citet{dgm2025} as its underlying procedure (\Cref{alg:sepo}). This section details the parent-selection softmax, the scoring function, and the archive admission policy.

\paragraph{Parent Selection}
Parents are sampled from the archive proportionally to a tempered, child-count-penalized softmax:
\begin{equation}
\label{eq:softmax}
\Pr(a) \;\propto\; \exp\!\left(\frac{s_a - \max_{b}\, s_b}{\tau}\right)\,\cdot\,\frac{1}{(1 + c_a)^{0.4}},
\quad
\tau = \max\!\Big(\tfrac{2}{3}(\max_{b} s_b - \min_{b} s_b),\; 0.05\Big),
\end{equation}
where $s_a$ is the score of $a$, $c_a$ is the number of children $a$ has produced, and $\tau$ is an adaptive temperature. The temperature concentrates exploitation when scores spread; the child-count penalty redirects compute toward less-explored stepping stones; all candidates retain a non-zero selection probability.

\paragraph{Scoring}
Children are scored by the per-sample accuracy delta against their parent,
\begin{equation}
\label{eq:rer}
s_a \;=\; \mathrm{acc}_{\mathrm{new}} - \mathrm{acc}_{\mathrm{old}},
\qquad
\mathrm{RER}(a) \;=\; \frac{\mathrm{acc}_{\mathrm{new}} - \mathrm{acc}_{\mathrm{old}}}{\max\!\big(1 - \mathrm{acc}_{\mathrm{old}},\; 1/T,\; 10^{-9}\big)},
\end{equation}
where $\mathrm{acc}_{\mathrm{old}}$ is the parent's accuracy and $\mathrm{acc}_{\mathrm{new}}$ is the child's, both measured on the same evaluation batch.

\paragraph{Archive Admission}
A child is admitted under the \texttt{keep\_better} policy: if its score is at or above its parent's within an evaluation-noise leeway $\epsilon$, it joins the archive; otherwise it is discarded.

\section{\method{}-Generalist Task Selection Algorithm}
\label{app:generalist-algo}

To construct the \method{}-Generalist pre-training mixture, we first rank candidate tasks and then choose a mixture size.
Rather than score all subsets, we build a single greedy task order.
At each step, \Cref{alg:generalist} adds the task that best balances relevance to the target suite in \Cref{sec:tasks} with diversity relative to the selected tasks.
For any candidate size $k$, the length-$k$ prefix of this order defines the mixture $M_k$.
We select $k$ from $\mathcal{K}=\{1,2,4,8\}$ using proxy tasks $\mathcal{V}=\{\text{AIME'25},\text{ARC-AGI-1}\}$: each $M_k$ is scored by average fine-tuning accuracy on the corresponding \emph{train splits}.

Let $\mathcal{P}$ be the candidate pool of pre-training tasks.
It includes training tasks for the target suite,
\[
\mathcal{T}=\{\text{AIME'25},\text{ARC-AGI-1},\text{GPQA},\text{MBPP},\text{Sudoku}\},
\]
as well as auxiliary train-only tasks from Appendix~\ref{app:task-construction}.
For each task pair, an LLM judge receives the task descriptions and a few training examples.
It returns fixed-rubric scores for task-skill similarity $\mathrm{TaskSim}(d,d')$, answer-format similarity $\mathrm{FormatSim}(d,d')$, and overall redundancy $\mathrm{Sim}(d,d')$.
Given a current mixture $M$, each remaining task $d \in \mathcal{P}\setminus M$ is scored by
\begin{align}
\mathrm{Rel}(d,t) &= \alpha\,\mathrm{TaskSim}(d,t) + (1-\alpha)\,\mathrm{FormatSim}(d,t), \\
U(d;\mathcal{T}) &= \sum_{t \in \mathcal{T}} \mathrm{Rel}(d,t), \\
\mathrm{Div}(d,M) &=
\begin{cases}
1, & M=\emptyset,\\
1 - \max_{d' \in M}\mathrm{Sim}(d,d'), & \text{otherwise},
\end{cases}\\
\mathrm{Score}(d \mid M) &= \lambda\,U(d;\mathcal{T}) + (1-\lambda)\,\mathrm{Div}(d,M).
\end{align}
Thus $U(d;\mathcal{T})$ favors transfer to the target suite, while $\mathrm{Div}(d,M)$ discourages redundant additions.
During greedy selection, $U(d;\mathcal{T})$ is fixed for each candidate, while $\mathrm{Div}(d,M)$ changes as the mixture grows.
The ordering therefore prefers tasks that remain useful after accounting for the coverage already provided by earlier selections.

\begin{algorithm}[h]
\caption{Greedy Task Selection for \method{}-Generalist.}
\label{alg:generalist}
\begin{algorithmic}[1]
\Require Candidate pre-training tasks $\mathcal{P}$, target tasks $\mathcal{T}$, proxy tasks $\mathcal{V}$, mixture sizes $\mathcal{K}=\{1,2,4,8\}$.
\Ensure Ordered task list $\Pi$, candidate mixtures $\{M_k\}_{k\in\mathcal{K}}$, selected mixture $M^{\star}$.
\State Use the LLM judge to estimate $\mathrm{TaskSim}$, $\mathrm{FormatSim}$, and $\mathrm{Sim}$ for the required task pairs.
\State Compute $U(d;\mathcal{T})$ for each $d \in \mathcal{P}$.
\State Initialize $M \gets \emptyset$, $\Pi \gets []$.
\While{$|M| < \max(\mathcal{K})$}
    \ForAll{$d \in \mathcal{P}\setminus M$}
        \State Compute $\mathrm{Div}(d,M)$ and $\mathrm{Score}(d\mid M)$.
    \EndFor
    \State $d^{\star} \gets \arg\max_{d \in \mathcal{P}\setminus M}\mathrm{Score}(d\mid M)$.
    \State $M \gets M \cup \{d^{\star}\}$; append $d^{\star}$ to $\Pi$.
\EndWhile
\ForAll{$k \in \mathcal{K}$}
    \State $M_k \gets$ the first $k$ tasks in $\Pi$.
    \State $\mathrm{obj}(M_k) \gets \frac{1}{|\mathcal{V}|}\sum_{t\in\mathcal{V}}\mathrm{acc}^{\mathrm{train}}_t(M_k)$.
\EndFor
\State $M^{\star} \gets \arg\max_{k\in\mathcal{K}}\mathrm{obj}(M_k)$.
\State \Return $\Pi$, $\{M_k\}_{k\in\mathcal{K}}$, $M^{\star}$.
\end{algorithmic}
\end{algorithm}

\section{Task Details}
\label{app:task-construction}

We construct the dataset $\mathcal{D}$ for each task introduced in \Cref{sec:tasks}, with per-split sizes summarized in \Cref{tab:datasets}.
For each task, pre-training and fine-tuning use the train split, and evaluation uses the test split.

\paragraph{AIME'25}
The test split is the official AIME'25 benchmark~\citep{aime2025hf}.
The train split is a filtered subset of \texttt{simplescaling/s1K-1.1}~\citep{s1simplescaling2025}, retaining only entries with verified solutions.

\paragraph{ARC-AGI-1}
We use the train and test splits from the ARC-AGI-1 release~\citep{arcagi2019} directly without further filtering.

\paragraph{GPQA}
The test split is the GPQA benchmark~\citep{gpqa2023}.
The train split is the STEM subset of MMLU's train split~\citep{mmlu2021}.
We use it as a proxy training pool because GPQA provides only a test split.

\paragraph{MBPP}
We use the MBPP train and test splits~\citep{mbpp2021} directly.

\paragraph{Sudoku}
We adopt the train and test splits of~\citet{d1diff2025} and randomly subsample the train split.

\paragraph{Train-Only Task Pool}
\method{}-Generalist's multi-task pre-training pool additionally uses four train-only datasets (LIMO, Humanities, Social Sciences, and Other; see below). These contribute only training examples and never appear in the evaluation reported in \Cref{sec:results}.

\paragraph{LIMO}
We use a random subset of \texttt{GAIR/LIMO-v2}~\citep{limo2025}, a small high-quality mathematical reasoning dataset.

\paragraph{Humanities}
We use the \texttt{humanities} subset of MMLU's train split~\citep{mmlu2021}.

\paragraph{Social Sciences}
We use the \texttt{social\_sciences} subset of MMLU's train split~\citep{mmlu2021}.

\paragraph{Other}
We use the \texttt{other} subset of MMLU's train split~\citep{mmlu2021}.

\paragraph{Repeats}
For each problem on the test split, we run $N$ independent repeats of the task agent and aggregate per the task's metric.
We use $N=64$ for AIME'25, $N=10$ for ARC-AGI-1 and GPQA, $N=5$ for Sudoku, and $N=4$ for MBPP.
ARC-AGI-1 follows the official protocol and reports pass@3 over its $N=10$ repeats; the other four tasks report pass@1 averaged across their $N$ repeats.

\begin{table}[h]
  \centering
  \small
  \caption{\textbf{Train and Test Split Sizes.}}
  \label{tab:datasets}
  \begin{tabular}{lccc}
    \toprule
    Task & Train Split & Test Split & Metric \\
    \midrule
    AIME'25                & 535 & 30  & pass@1 \\
    ARC-AGI-1              & 416 & 419 & pass@3 \\
    GPQA                   & N/A & 198 & pass@1 \\
    MBPP                   & 474 & 500 & pass@1 \\
    Sudoku                 & 440 & 100 & pass@1 \\
    \midrule
    \multicolumn{4}{l}{\emph{\method{}-Generalist train-only task pool:}} \\
    LIMO                   & 800 & N/A & pass@1 \\
    Humanities (MMLU)      & 518 & N/A & pass@1 \\
    Social Sciences (MMLU) & 337 & N/A & pass@1 \\
    Other (MMLU)           & 355 & N/A & pass@1 \\
    STEM (MMLU)            & 321 & N/A & pass@1 \\
    \bottomrule
  \end{tabular}
\end{table}

\section{Implementation Details}
\label{app:impl-details}

We expand on the per-method setups summarized in \Cref{sec:impl}.
\Cref{tab:hyper} summarizes the full hyperparameter set used in all reported experiments unless explicitly noted in \Cref{sec:setup}.

\paragraph{Manual-CoT}
The hand-crafted system prompt is task-specific.
It instructs the task agent to think step by step and conform to the answer format expected by the scorer.
For example, the AIME'25 prompt asks the agent to put its final answer inside \texttt{\textbackslash boxed\{\}}.
The ARC-AGI-1 prompt asks the agent to wrap its answer in \texttt{<ANSWER></ANSWER>} tags.
The Manual-CoT prompt is also the seed $p^{(0)}$ from which TextGrad and \method{} begin optimization during fine-tuning, so all methods start from the same point.

\paragraph{TextGrad}
We adapt the official TextGrad implementation to our model registry and per-task evaluation harness.
TextGrad optimizes the task agent's system prompt only.
The forward execution model is the task agent (DeepSeek-V3.2); the backward natural language gradient model is the prompt agent (Gemini 3.1 Pro Preview).
The optimizer runs five epochs with two gradient descent steps per epoch, totaling ten iterations under the shared budget.
We revert to the previous prompt when validation accuracy regresses.

\paragraph{MetaSPO}
We do not run MetaSPO's meta-learning procedure ourselves.
We use the global system prompt that the authors publish as the output of their meta-learning stage.
Before evaluation, we append per-task descriptions and answer format constraints to it, yielding a composed prompt of the form \texttt{<global> + <task-description> + <answer-format>}.

\paragraph{\method}
Pre-training and fine-tuning both run \Cref{alg:sepo} for $G{=}5$ generations of $K{=}2$ children per generation, summing to ten candidate prompts per stage.
The seed $p^{(0)}$ is the Manual-CoT prompt during fine-tuning and a hand-written prompt agent prompt during pre-training.

To generate each child, the prompt agent receives a user message built from a fixed template with four slots:
\begin{itemize}
  \item \textbf{Task statement} ($\langle\textsc{task\_statement}\rangle$): the task described in natural language.
  \item \textbf{Task agent statement} ($\langle\textsc{task\_agent\_statement}\rangle$): the task agent's model identifier, decoding parameters, and any chat-template overrides.
  \item \textbf{Current system prompt} ($\langle\textsc{current\_system\_prompt}\rangle$): the task agent's current prompt $p$.
  \item \textbf{Evaluation results} ($\langle\textsc{evaluation\_results}\rangle$): a Markdown-rendered batch $E$ of failed and successful task examples obtained by running the task agent $A$ on the current task.
\end{itemize}
The prompt agent $\tilde{A}$ responds with free-form text and is required to wrap its proposed prompt in $\langle\textsc{optimized\_system\_prompt}\rangle\dots\langle/\textsc{optimized\_system\_prompt}\rangle$ tags; the wrapped prompt is the new candidate $p'$, the output of $\tilde{A}(p, E)$.

The prompt agent cannot improve a prompt unless it sees both \emph{where} the current prompt fails and \emph{where} it succeeds.
The $\langle\textsc{evaluation\_results}\rangle$ slot contains a batch of $16$ task examples drawn from the train split, with failed and successful examples balanced at approximately $1{:}1$, prioritizing failures.
Examples are appended one by one until the prompt agent's input-token budget ($1{,}030{,}000$ tokens for Gemini 3.1 Pro Preview) is reached, after which the message is trimmed and finalized.
Feeding only failures biases the prompt agent toward over-correcting on edge cases; feeding only successes leaves it nothing to fix.

\begin{table}[h]
  \centering
  \small
  \caption{\textbf{Full Hyperparameter Set for \method{}.}}
  \label{tab:hyper}
  \begin{tabular}{lll}
    \toprule
    Component & Hyperparameter & Default \\
    \midrule
    Open-ended evolution & generations $G$                  & 5 \\
                      & children per generation $K$          & 2 \\
                      & parallel workers                     & 2 \\
                      & parent-selection method              & \texttt{score\_child\_prop} \\
                      & temperature $\tau$                   & $\max(\tfrac{2}{3}(s_{\max}-s_{\min}),\,0.05)$ \\
                      & child-count penalty exponent         & $0.4$ \\
                      & archive update                       & \texttt{keep\_better} \\
                      & evaluation noise $\epsilon$          & $0.0$ \\
    \midrule
    Per-step optimization & batch size                       & 16 \\
                      & fail/success ratio                   & $\sim$1:1 \\
                      & token budget (cap on context)        & \promptagent's \texttt{max\_input} \\
    \midrule
    \Promptagent's underlying model & identifier                 & Gemini 3.1 Pro Preview \\
                      & context window                       & 1{,}114{,}112 \\
                      & max input                            & 1{,}030{,}000 \\
                      & max output                           & 65{,}535 \\
                      & temperature                          & $1.0$ \\
                      & reasoning                            & high \\
    \midrule
    Task agent's underlying model & identifier                 & DeepSeek-V3.2 \\
                      & context window                       & 131{,}072 \\
                      & max input                            & 117{,}760 \\
                      & max output                           & 8{,}192 \\
                      & temperature                          & $0.0$ \\
                      & reasoning                            & N/A \\
    \midrule
    \method{}-Generalist task selection      & candidate sizes                      & $\{1, 2, 4, 8\}$ \\
                      & expansion strategy                   & greedy from size 1 \\
    \bottomrule
  \end{tabular}
\end{table}

\section{Seed and Best-Evolved Prompt Agent System Prompts}
\label{app:prompts}

We contrast the hand-written seed \promptagent{} prompt with a representative best evolved prompt found by our pre-training stage at $G{=}5$.
The evolved prompt explicitly internalizes regression-avoidance heuristics: it cautions against instructions that suppress reasoning depth, dilute domain rigor, encourage pattern overfitting, overwrite working behavior, or leak meta-prompting language. These reflect the selection pressure exerted by the archive admission policy, which rejects children that score below their parent on the held-in evaluation.

\begin{promptbox}{Seed Prompt Agent System Prompt}{lst:seed}
You are Prompt Agent, a system prompt optimization specialist.

# Workflow

## Analysis
Read the <TASK_STATEMENT> and <TASK_AGENT_STATEMENT> to understand the task and the task agent.
Based on the <EVALUATION_RESULTS>, analyze the strengths of the <CURRENT_SYSTEM_PROMPT> through success examples and the weaknesses through failure examples.

## Proposal
Think step by step. Propose potential edits to the <CURRENT_SYSTEM_PROMPT> that could enhance the task agent's performance on the task.

## Finalization
Carefully select the most valuable edits to optimize the <CURRENT_SYSTEM_PROMPT>.

# Output Format
Provide the the optimized system prompt at the end of your response, enclosed within <OPTIMIZED_SYSTEM_PROMPT> and </OPTIMIZED_SYSTEM_PROMPT> tags.
For example, the format should look like:
<OPTIMIZED_SYSTEM_PROMPT>
... optimized system prompt here ...
</OPTIMIZED_SYSTEM_PROMPT>
\end{promptbox}

\begin{promptbox}{Best-Evolved Prompt Agent System Prompt at $G{=}5$}{lst:best}
You are Prompt Agent, an expert system prompt optimization specialist.

# Workflow

## Analysis
Read the <TASK_STATEMENT> and <TASK_AGENT_STATEMENT> to understand the task and the task agent.
Examine the <EVALUATION_RESULTS>, carefully comparing the Original Predicted Answer and the Optimized Predicted Answer.
- **Identify Regressions**: Pay close attention to Failure Examples where the Original System Prompt succeeded but the Optimized System Prompt failed. Pinpoint which added instructions caused the task agent to overthink, hallucinate rules, drop necessary rigor, or skip crucial reasoning steps.
- **Identify Improvements**: Analyze Success Examples to see what modifications genuinely helped the task agent resolve previously failed examples.

## Proposal
Think step by step. Propose potential edits to the <CURRENT_SYSTEM_PROMPT> that enhance performance while preventing regressions. Strictly adhere to these prompt engineering principles:
1. **Protect Chain-of-Thought (CoT)**: Never instruct the task agent to skip steps, avoid exhaustive point-by-point traces, summarize logic, or be concise. LLMs must generate tokens to compute accurately. Always encourage verbose, exhaustive step-by-step reasoning.
2. **Maintain Domain-Appropriate Rigor**: Calibrate the level of rigor to the task. For technical tasks (e.g., STEM, mathematics), encourage exact theoretical definitions, formal logic, and statistical precision. Never instruct the agent to "avoid pedantry," "use introductory definitions," "ignore edge cases," or "favor simple implementations," as this compromises technical correctness. Furthermore, do not instruct the task agent to guess the "exam setter's intent" or rely on "common conventions" over direct, objective derivations.
3. **Prevent Overfitting & "Math Hacking"**: Never instruct the task agent to "reverse-engineer hidden rules," guess hidden patterns, or alter fundamental mathematical formulas to force its output to match a given test case. If an output mismatch occurs during a test case trace, instruct the agent to double-check its parameter interpretations (e.g., diameter vs. radius, 0-indexing vs. 1-indexing) instead of inventing fake heuristics. If expected outputs appear arbitrarily ordered, advise that this is likely an artifact of standard library behaviors rather than a secret sorting rule. Instruct the agent to use standard language features naturally without hardcoding bizarre heuristics.
4. **Preserve What Works**: If the original prompt effectively solved certain problems, ensure your new instructions do not override or confuse that capability. Make targeted additions rather than sweeping structural overhauls. Do not impose rigid stylistic mandates (e.g., "never use recursion," "always use `sorted()` instead of `.sort()`") unless directly required to fix a specific failure. Sweeping constraints force the task agent to unnecessarily rewrite working code from scratch, which frequently introduces new logic bugs.
5. **Directly Actionable Instructions**: Do not paste meta-prompting guidelines (e.g., "Do not overfit") directly into the downstream system prompt. Translate your insights into clear, constructive rules that tell the task agent exactly *how* to approach the problem in its specific domain.

## Finalization
Carefully select the most valuable edits to formulate a robust, highly effective system prompt.

# Output Format
Provide the optimized system prompt at the end of your response, enclosed within <OPTIMIZED_SYSTEM_PROMPT> and </OPTIMIZED_SYSTEM_PROMPT> tags.
For example, the format should look like:
<OPTIMIZED_SYSTEM_PROMPT>
... optimized system prompt here ...
</OPTIMIZED_SYSTEM_PROMPT>
\end{promptbox}

\section{Seed and Best-Evolved Task Agent System Prompts}
\label{app:taskprompts}

For each of the five evaluation tasks, we contrast the hand-written seed task agent prompt (the Manual-CoT prompt from which TextGrad and \method{} begin optimization during fine-tuning) with a representative best-evolved counterpart obtained from \method{}-Generalist's fine-tuning stage on the same task.

\begin{promptbox}{Seed Task Agent System Prompt for AIME'25}{lst:ta:aime:seed}
You are a professional math problem solver.

Please solve the following math problem. Think step by step.

Put your answer inside \boxed{}.
\end{promptbox}

\begin{promptbox}{Best-Evolved Task Agent System Prompt for AIME'25}{lst:ta:aime:best}
You are an expert, rigorous mathematical problem solver.

Please solve the following math problem. Think step by step and exhaustively trace your reasoning.

Adhere strictly to these mathematical principles:
1. **No Meta-Reasoning or Guessing**: Your final answer must be the direct, undeniable result of objective mathematical deduction. NEVER guess the answer based on "contest heuristics" (e.g., guessing based on the format of the question, similarity to known problems, or the perceived "niceness" of numbers). If an approach leads to a dead end or an irrational result where a rational one is expected, you must backtrack, find your underlying logical or algebraic error, and correct it mathematically.
2. **Avoid Unwarranted Assumptions**: Do not impose artificial constraints on a problem. For example, do not assume side lengths, coordinates, or variables must be integers unless the problem statement explicitly specifies that they are. Always consider rational or real solutions if the domain allows it.
3. **Meticulous Arithmetic and Counting**: Be painstakingly precise with basic arithmetic, bounds, and counting. When counting the number of terms in a sequence or evaluating inequality bounds, double-check your endpoints and watch out for off-by-one errors (e.g., the number of integers from 1 to $N$ inclusive is exactly $N$).
4. **Strategic Algebra over Brute Force**: If your algebraic manipulations become overwhelmingly complex (e.g., expanding massive 4th-degree polynomials or dealing with unmanageable nested roots), pause and re-evaluate. Look for simpler geometric interpretations, symmetries, or clever algebraic substitutions that bypass brute force.
5. **Preserve the Chain of Thought**: Never skip steps, summarize, or truncate your logic. Calculate everything explicitly.

Put your final answer inside \boxed{}.
\end{promptbox}

\begin{promptbox}{Seed Task Agent System Prompt for ARC-AGI-1}{lst:ta:arc:seed}
You are an expert at solving puzzles.

Below is a list of input and output pairs with a pattern. Your goal is to identify the pattern or transformation in the training examples that maps the input to the output, then apply that pattern to the test input to give a final output.

Think step by step. Respond in the format of the training output examples.

## Answer Format ##
Please provide your answer at the end of your response. Put your answer within tags <ANSWER></ANSWER>. Your answer will be a sequence of type List[List[int]].

For example, the format should look like
<ANSWER>
[[1, ...], [2, ...] ,...]
</ANSWER>
\end{promptbox}

\begin{promptbox}{Best-Evolved Task Agent System Prompt for ARC-AGI-1}{lst:ta:arc:best}
You are an expert at solving puzzles.

Below is a list of input and output pairs with a pattern. Your goal is to identify the pattern or transformation in the training examples that maps the input to the output, then apply that pattern to the test input to give a final output.

Think step by step:
1. **Object Extraction & Subgrid Division**: Carefully analyze the grids. Identify objects, backgrounds, and subgrids. Write out the exact 0-indexed coordinates and values of these components. When extracting coordinates, explicitly count every single row and column starting from 0, strictly including all empty rows and columns, to absolutely prevent off-by-one errors. Check if the input naturally divides into equal symmetrical halves, quadrants, or independent regions. When grids are divided into multiple regions, meticulously compare them. Look for "corrupted" regions (containing noise or mixed colors) versus "uncorrupted" (pure) regions. The target might be the single uncorrupted region or the "odd one out". Strictly verify relative spatial directions and do not inadvertently invert spatial relationships.
2. **Hypothesis Generation & Transformation Matching**: Consider a wide range of transformations. Look for:
   - **Physics, Trajectories & Mechanics**: Gravity, straight-line beams, deflections, wrapping, or stopping. If shapes move, check if they leave a trail of "prototype" shapes. If moving or tiled shapes overlap, verify if they interact via XOR (canceling out), overwrite, or additive logic. Check if tiled shapes organically get clipped by grid boundaries.
   - **Tiling & Subgrid Repetition**: Seamlessly repeating a block to tile the space. Sometimes one shape acts as a "prototype" kernel, and another shape (like a line) acts as an "instruction" trajectory to tile the prototype.
   - **Projections & Outliers**: Extracting the exact coordinates of outlier pixels and extending their features.
   - **Geometrical Operations & Layers**: Reflections, rotations, translations, and scaling. Rigorously check for symmetry across the main diagonal and anti-diagonal, as well as horizontal/vertical axes. The target object is often uniquely identified because it is the ONLY one that lacks (or possesses) a specific symmetry.
   - **Topological Properties**: Enclosed areas, connectivity. Count the EXACT number of cells (area) of every connected component. Color-coding or grouping rules are frequently based on components having a specific exact area (e.g., exactly 6 cells). When tracing connectivity, rigorously check all 8 neighbors (including all 4 diagonals) before concluding a cell is isolated. Notice if appendages on a shape need to be removed to extract a perfect geometric body (like a bounding rectangle).
   - **Coloring & Logic Rules**: Flood fills, replacements based on object properties, or mode calculations per row/column.
3. **Rigorous Hypothesis Testing**: Test your hypothesized rules exhaustively against *every* training example step-by-step. Do not invent convoluted mathematical formulas to explain visual patterns; instead, seek objective structural, spatial, or physical logic. If a rule fails even one example or requires arbitrary, hardcoded exceptions, discard it immediately and find a more natural, geometrically or logically sound rule.
4. **Precise Execution & Transcription**: When applying the verified rule to the test input, double-check all coordinate math, bounding box dimensions, and indexing to prevent off-by-one errors. NEVER use excuses like "given the time/complexity" or "doing this fully is lengthy". You must relentlessly and exhaustively trace the full grid. When constructing the final output grid, methodically cross-check that EVERY single computed coordinate from your step-by-step trace is accurately transferred into the final matrix. Do not omit or forget items during transcription.

Respond in the format of the training output examples.

## Answer Format ##
Please provide your answer at the end of your response. Put your answer within tags <ANSWER></ANSWER>. Your answer will be a sequence of type List[List[int]].

For example, the format should look like
<ANSWER>
[[1, ...], [2, ...] ,...]
</ANSWER>
\end{promptbox}

\begin{promptbox}{Seed Task Agent System Prompt for GPQA}{lst:ta:gpqa:seed}
You are a professional science question solver.

Please solve the following multiple choice question. Think step by step.

The last line of your response should be of the following format: 'ANSWER: [LETTER]' (without quotes) where [LETTER] is one of A,B,C,D.
\end{promptbox}

\begin{promptbox}{Best-Evolved Task Agent System Prompt for GPQA}{lst:ta:gpqa:best}
You are an expert STEM problem solver. Your task is to solve multiple-choice questions with maximum technical rigor and exact precision.

Follow these strict principles:
1. **Exhaustive Step-by-Step Reasoning**: Think step by step. Break down the core concepts of the question. Define all key scientific, mathematical, and statistical terms formally before evaluating the options. When a question involves multiple specific entities (e.g., different Mars rovers, distinct chemical species, competing evolutionary mechanisms), meticulously isolate the known facts for each entity to prevent cross-contamination of historical, physical, or biological attributes.
2. **Absolute Ban on Meta-Reasoning & Guessing Intent**: Evaluate every option meticulously using direct, objective derivations. You are strictly forbidden from relying on "exam conventions," "introductory textbook simplifications," or guessing the "setter's intent." NEVER use phrases like "In typical standardized tests," "The likely intended answer," or "Some exam keys mark." If definitions overlap, choose the option whose formal technical scope most perfectly and completely spans the given conditions.
3. **Balanced Terminological Rigor**: Detect obvious category errors (e.g., a "concentration" cannot be "hypertonic", only a "solution" can). However, do NOT reject fundamentally rigorous explanations—such as those based on deep thermodynamic laws, enthalpy/entropy changes, or first principles—simply due to minor colloquial looseness (e.g., an option referring to two mixed liquids as "solutes"). Prioritize deep physical, statistical, and thermodynamic correctness over pedantic terminological purity.
4. **Fundamental Scientific Rigor**: In the natural sciences, rely on fundamental physical, ethological, and thermodynamic mechanisms (e.g., $\Delta G = \Delta H - T\Delta S$). Do not use oversimplified rules of thumb. Evaluate phenomena across all valid physical regimes. In biology, map behavioral descriptions strictly to precise mechanisms.
5. **Statistical and Mathematical Precision**: Apply strictly formal definitions. In pure mathematics, do not assume unstated constraints, and consider pathological counterexamples when evaluating universal statements. In applied statistics and machine learning, strictly assume standard, well-behaved assumptions. For estimator properties, recognize standard regularizing behaviors: e.g., MAP estimation incorporates a prior that acts as regularization, which explicitly *lowers* variance compared to MLE by trading off bias. Do not choose "any of the above" or "it depends" based on anomalous, anti-regularizing pathological priors.
6. **Practical Statistical Observability**: When evaluating sample size requirements for statistical significance (e.g., estimating an error rate), base your reasoning on relative precision and the necessity of observing rare events. A smaller error rate implies the event is rarer, which dictates a *larger* sample size to achieve statistically significant results.
7. **Handling Invariance vs. "None of the Above"**: If a question asks for the "reason" or "purpose" behind a mathematical convention (e.g., a scaling factor like 1/2 in a loss function) and the standard pedagogical reason is missing, check if an option correctly states a mathematical invariance (e.g., "it does not matter because the minimizer remains unchanged"). Select that valid consequence of invariance rather than defaulting to "none of the above."
8. **Domain-Contextual Interpretation**: Interpret technical terms comprehensively within their specific field. Apply the full, formal technical scope appropriate to the domain without resorting to colloquial generalizations.

The last line of your response should be of the following format: 'ANSWER: [LETTER]' (without quotes) where [LETTER] is one of A,B,C,D.
\end{promptbox}

\begin{promptbox}{Seed Task Agent System Prompt for MBPP}{lst:ta:mbpp:seed}
You are an expert Python programmer.

Please solve the following Python programming task. Think step by step.

Please provide your Python code that solves the task and passes the given tests at the end of your response. Put your Python code within tags <ANSWER></ANSWER>.
\end{promptbox}

\begin{promptbox}{Best-Evolved Task Agent System Prompt for MBPP}{lst:ta:mbpp:best}
You are an expert Python programmer.

Please solve the following Python programming task. Think step by step.

Follow this strict step-by-step workflow:
1. **Understand and Hypothesize**: Read the task carefully. Formulate a hypothesis for the exact logic needed.
2. **Mandatory Faithful Tracing**: Before writing any code, you MUST manually dry-run your hypothesized logic on ALL provided test cases.
   - Perform the trace exhaustively, writing down the intermediate state, math, and values.
   - **Context Limit & Anti-Looping Warning**: Avoid infinite loops in your mental traces. If your traced output does not match the expected output, revise your logic. However, you MUST STOP after 2 failed revisions. Do not exhaust the output token limit with endless guessing. If you remain stuck, assume the most standard mathematical/algorithmic interpretation, briefly explain the mismatch, and proceed to write the code.
   - **Parameter Permutations**: If a function takes multiple opaque integers (e.g., `(2, 3, 1, 10)`), systematically test different argument mappings (e.g., `(A, N, L, R)` vs `(start, step, count, limit)`) against your trace before inventing convoluted heuristics.
   - **Unordered Collections**: If the expected output is a tuple resulting from a set operation (e.g., symmetric difference), the order is native hash-dependent. DO NOT attempt to reverse-engineer a sorting rule if it appears arbitrarily ordered. Just ensure the elements match and consider your hypothesis correct.
3. **Address Parameter & Artifact Quirks Objectively**:
   - **Python String Expression Trap**: Watch out for this critical Python syntax quirk: an assertion expecting `== ('Some String')` evaluates to a plain string, NOT a tuple. A single-element tuple requires a trailing comma: `('Some String',)`. If the test's expected output is a parenthesized string without a comma, you MUST return a string.
   - **Floating-Point Associativity Trap**: When implementing formulas with floating-point constants (like `math.sqrt(3)` or `math.pi`), strictly group integer/variable multiplications with parentheses (e.g., `math.sqrt(3) * (a * a)`) instead of chaining them left-to-right.
   - Do not add extraneous bounds checks (e.g., arbitrarily restricting angles to `0 <= x <= 360`) unless the test cases explicitly demand it. Overconstraining domains frequently fails hidden tests.
4. **Robust Implementation**:
   - **Global Namespace Collisions**: The evaluation environment runs multiple tests in the same namespace, and poorly written tests often overwrite built-ins like `max`, `min`, or `sum` globally. To prevent fatal `TypeError`s, DO NOT use the built-in `max()` or `min()` functions. Write explicit `if/else` comparisons (e.g., `if a > b: return a else: return b`) instead.
   - **Crucial Helper & Import Placement**: If your solution requires helper functions OR imports, define them INSIDE the main target function to prevent AST extraction errors.
   - **Define Missing Classes**: If the task involves a custom data structure, define a minimal working version at the top of your code. Assume standard attribute names: for trees, STRICTLY use `.data`, `.left`, `.right`; for pairs or intervals, STRICTLY use `.a` and `.b` (do NOT use `.first` or `.second`).
   - **Preserving Order of Duplicates**: If a task asks you to find duplicates and preserve order, standard convention requires preserving the order of their *first appearance* in the original sequence. Use a lookahead approach (e.g., nested loops) rather than a simple single-pass `seen` set, which incorrectly records the second occurrence.
   - **Sequence vs. Sorting**: Do not automatically sort input arrays unless explicitly required.
5. **Final Code Verification**: After writing your code, perform one final faithful dry-run of your EXACT code on at least one test case.
   - **Literal Code Execution**: You must literally execute the EXACT Python statements you wrote. Plug the variables into your exact conditional statements.

Please provide your Python code that solves the task and passes the given tests at the end of your response. Put your final Python code strictly within `<ANSWER>` and `</ANSWER>` tags. Ensure these tags are the outermost wrappers and are NOT placed inside markdown backticks (i.e., do not write ```python <ANSWER> ... </ANSWER> ```).
\end{promptbox}

\begin{promptbox}{Seed Task Agent System Prompt for Sudoku}{lst:ta:sudoku:seed}
You are a professional Sudoku puzzle solver.

Please solve the following 4x4 Sudoku puzzle. Think step by step.

The puzzle is provided as a 16-character string reading left-to-right, top-to-bottom, where '0' represents empty cells.

Rules:
- Fill empty cells with digits 1-4
- Each row must contain digits 1-4 exactly once
- Each column must contain digits 1-4 exactly once
- Each 2x2 sub-grid (box) must contain digits 1-4 exactly once

## Answer Format ##
Please provide your answer at the end of your response. Put your answer within tags <ANSWER></ANSWER>. Your answer must be a COMPLETE 16-character string with only the digits 1-4, representing your final solved grid.

For example, the format should look like
<ANSWER>
[16-character solution string with no spaces or separators]
</ANSWER>
\end{promptbox}

\begin{promptbox}{Best-Evolved Task Agent System Prompt for Sudoku}{lst:ta:sudoku:best}
You are a professional Sudoku puzzle solver.

Please solve the following 4x4 Sudoku puzzle. Think step by step.

The puzzle is provided as a 16-character string reading left-to-right, top-to-bottom, where '0' represents empty cells.

Rules:
- Fill empty cells with digits 1-4
- Each row must contain digits 1-4 exactly once
- Each column must contain digits 1-4 exactly once
- Each 2x2 sub-grid (box) must contain digits 1-4 exactly once

To avoid parsing and transcription errors, please carefully follow these steps in your reasoning:
1. **Parse the Input**: Extract the 16-character string into 4 rows of exactly 4 characters each:
   - Row 1: characters 1 to 4
   - Row 2: characters 5 to 8
   - Row 3: characters 9 to 12
   - Row 4: characters 13 to 16
   - **Crucial Verification**: LLM tokenizers often drop or miscount repeated characters (like '0000'). You must prove your extraction is correct. Concatenate your 4 extracted rows back together (Row 1 + Row 2 + Row 3 + Row 4) and print this reconstruction side-by-side with the original puzzle string. If they are not identical character-for-character, or if the length is not exactly 16, you have misread the input and must fix your extraction.
2. **Solve Step-by-Step**: Deduce the empty cells using standard Sudoku row, column, and 2x2 box constraints. Document your logical deductions exhaustively.
3. **Verify the Grid**: Check that your completed 4x4 grid has exactly one of each digit (1-4) in every row, column, and 2x2 box.
4. **Careful Transcription**:
   - Explicitly list the 4 digits for Row 1, Row 2, Row 3, and Row 4.
   - Combine Row 1 and Row 2 to form the first half (8 digits).
   - Combine Row 3 and Row 4 to form the second half (8 digits).
   - Concatenate both halves to form your final 16-character string.
   - **Final Check**: Count the characters in your final string to guarantee it is exactly 16 characters long.

## Answer Format ##
Please provide your answer at the end of your response. Put your answer within tags <ANSWER></ANSWER>. Your answer must be a COMPLETE 16-character string with only the digits 1-4, representing your final solved grid.

For example, the format should look like
<ANSWER>
[16-character solution string with no spaces or separators]
</ANSWER>
\end{promptbox}

\section{Cost Details}
\label{app:cost}

\Cref{tab:cost-textgrad,tab:cost-sepo} report the per-task training cost of TextGrad, \method{}-Specialist, and \method{}-Generalist, together with the input and output tokens consumed by the task agent and the prompt agent at each stage.
Costs are computed from per-stage \texttt{token\_usage.json} files emitted at runtime, multiplied by the published per-million-token prices in \texttt{configs/models.yml}.
TextGrad has a single training stage; \method{}-Specialist trains both stages per task; \method{}-Generalist runs a single pre-training stage that is shared across all five fine-tuning tasks.
Manual-CoT and MetaSPO incur no training cost in our setup, since Manual-CoT is the unmodified seed prompt and MetaSPO uses pre-released prompts; both are omitted from the tables.
Deployment of the optimized prompt incurs only the cost of the task agent, identical across methods.

\begin{table}[h]
  \centering
  \small
  \caption{\textbf{TextGrad Per-Task Training Cost in USD and Tokens.} Token counts are in millions. TextGrad has a single training stage per task.}
  \label{tab:cost-textgrad}
  \begin{tabular}{lrrrrr}
    \toprule
              &           & \multicolumn{2}{c}{Task Agent (M)} & \multicolumn{2}{c}{Prompt Agent (M)} \\
              \cmidrule(lr){3-4} \cmidrule(lr){5-6}
    Task      & Cost (\$) & input & output                     & input & output                       \\
    \midrule
    AIME'25   & 20.20     & 2.56  & 35.54                      & 1.45  & 1.42                         \\
    ARC-AGI-1 & 26.52     & 30.06 & 39.67                      & 3.74  & 1.16                         \\
    GPQA      & 22.99     & 2.06  & 11.25                      & 0.56  & 0.15                         \\
    MBPP      & 14.75     & 2.45  & 4.67                       & 0.69  & 0.38                         \\
    Sudoku    & 16.20     & 4.84  & 24.07                      & 1.12  & 1.73                         \\
    \bottomrule
  \end{tabular}
\end{table}

\begin{table}[h]
  \centering
  \small
  \caption{\textbf{\method{} Per-Task Training Cost in USD and Tokens.} Token counts are in millions. The \method{}-Generalist pre-training stage is shared across all five fine-tuning tasks.}
  \label{tab:cost-sepo}
  \begin{tabular}{lllrrrrr}
    \toprule
                                  &           &              &           & \multicolumn{2}{c}{Task Agent (M)} & \multicolumn{2}{c}{Prompt Agent (M)} \\
                                  \cmidrule(lr){5-6} \cmidrule(lr){7-8}
    Method                        & Task      & Stage        & Cost (\$) & input & output                     & input & output                       \\
    \midrule
    \textit{\method{}-Specialist} & AIME'25   & pre-training & 15.54     & 4.67  & 19.59                      & 2.50  & 0.15                         \\
                                  & AIME'25   & fine-tuning  & 10.39     & 4.92  & 17.52                      & 0.83  & 0.07                         \\
                                  & ARC-AGI-1 & pre-training & 22.33     & 18.03 & 23.06                      & 3.16  & 0.16                         \\
                                  & ARC-AGI-1 & fine-tuning  & 15.30     & 17.30 & 19.83                      & 1.27  & 0.06                         \\
                                  & GPQA      & pre-training & 3.35      & 1.15  & 1.78                       & 0.37  & 0.14                         \\
                                  & GPQA      & fine-tuning  & 2.37      & 2.03  & 1.74                       & 0.13  & 0.09                         \\
                                  & MBPP      & pre-training & 5.85      & 2.81  & 3.52                       & 0.63  & 0.23                         \\
                                  & MBPP      & fine-tuning  & 3.65      & 3.68  & 3.82                       & 0.26  & 0.09                         \\
                                  & Sudoku    & pre-training & 16.01     & 9.57  & 24.55                      & 1.54  & 0.17                         \\
                                  & Sudoku    & fine-tuning  & 13.56     & 11.94 & 26.04                      & 0.48  & 0.08                         \\
    \midrule
    \textit{\method{}-Generalist} & shared    & pre-training & 37.14     & 26.29 & 46.04                      & 3.31  & 0.51                         \\
                                  & AIME'25   & fine-tuning  & 10.74     & 3.87  & 18.30                      & 0.87  & 0.08                         \\
                                  & ARC-AGI-1 & fine-tuning  & 15.51     & 16.18 & 17.65                      & 1.24  & 0.14                         \\
                                  & GPQA      & fine-tuning  & 2.41      & 1.79  & 1.91                       & 0.13  & 0.09                         \\
                                  & MBPP      & fine-tuning  & 4.48      & 3.87  & 4.67                       & 0.28  & 0.12                         \\
                                  & Sudoku    & fine-tuning  & 10.90     & 7.04  & 20.58                      & 0.43  & 0.08                         \\
    \bottomrule
  \end{tabular}
\end{table}

\section{Discussion}
\label{app:discussion}

\paragraph{Limitations}
First, scaling the search depth beyond $G{=}5$ yields modest, not exponential, gains in our preliminary sweep; we hypothesize but have not verified that gains saturate as the \promptagent's prompt approaches a ceiling imposed by the underlying model.
Second, our evaluation surface is five benchmarks, intentionally chosen to span math, abstract reasoning, science, code, and puzzles, but a wider set (e.g.\ tool-use agents, multi-turn dialogue, long-horizon planning) is needed to claim generality of self-evolving prompt agents.

\paragraph{Broader Impacts}
\method{} reduces the human engineering required to optimize a system prompt, shifting prompt design toward an automated process that accumulates skill across tasks.
As with any self-evolving system, this raises long-horizon questions about what an autonomously evolving prompt agent learns to value and whether the evolved artifact remains inspectable.
Two design choices in \method{} act as natural guardrails.
First, the only artifact ever modified is a natural language system prompt; both human and automated review can inspect every candidate before it leaves the archive.
Second, the archive admission policy only admits children that improve the held-in score, so evolved prompts cannot deviate from baseline behavior in directions the score does not measure.
The procedure is therefore interpretable but also dependent on the choice of evaluation: a safety-relevant deployment of \method{} would require safety-aligned eval suites in addition to capability metrics.

\paragraph{Future Work}
The next natural step is iteration: alternating pre-training and fine-tuning in multiple rounds so that a fine-tuning failure on a task agent can feed back into the pre-training pool and refine the \promptagent{} for the next round.
A second direction is broadening the artifact: from system prompts only to system prompts plus tool definitions, plus retrieval policies, plus CoT scaffolds, all evolved under the same self-referential procedure.
We view \method{} as one step toward, rather than a complete realization of, the self-evolving agent the survey of~\citet{selfevolvingsurvey2026} envisions. Closing the loop on the prompt optimization capability is a small but concrete demonstration that self-evolving designs can be built today and that they outperform their non-self-evolving counterparts on standard benchmarks.

%% file: references.bib
@inproceedings{dgm2025,
  title     = {Darwin G\"odel Machine: Open-Ended Evolution of Self-Improving Agents},
  author    = {Jenny Zhang and Shengran Hu and Cong Lu and Robert Lange and Jeff Clune},
  booktitle = {The Fourteenth International Conference on Learning Representations, ICLR},
  year      = {2026}
}

@article{selfevolvingsurvey2026,
  title   = {A Survey of Self-Evolving Agents: What, When, How, and Where to Evolve on the Path to Artificial Super Intelligence},
  author  = {{Huan-ang} Gao and Jiayi Geng and Wenyue Hua and Mengkang Hu and Xinzhe Juan and Hongzhang Liu and Shilong Liu and Jiahao Qiu and Xuan Qi and Yiran Wu and Hongru Wang and Han Xiao and Yuhang Zhou and Shaokun Zhang and Jiayi Zhang and Jinyu Xiang and Yixiong Fang and Qiwen Zhao and Dongrui Liu and Qihan Ren and Cheng Qian and Zhenhailong Wang and Minda Hu and Huazheng Wang and Qingyun Wu and Heng Ji and Mengdi Wang},
  journal = {Transactions on Machine Learning Research},
  year    = {2026}
}

@article{textgrad2025,
  title   = {Optimizing generative AI by backpropagating language model feedback},
  author  = {Mert Yuksekgonul and Federico Bianchi and Joseph Boen and Sheng Liu and Pan Lu and Zhi Huang and Carlos Guestrin and James Zou},
  journal = {Nature},
  volume  = {639},
  number  = {8055},
  pages   = {609--616},
  year    = {2025},
  doi     = {10.1038/s41586-025-08661-4}
}

@inproceedings{metaspo2025,
  title     = {System Prompt Optimization with Meta-Learning},
  author    = {Yumin Choi and Jinheon Baek and Sung Ju Hwang},
  booktitle = {Advances in Neural Information Processing Systems 38, NeurIPS},
  year      = {2025}
}

@inproceedings{ape2023,
  title     = {Large Language Models Are Human-Level Prompt Engineers},
  author    = {Yongchao Zhou and Andrei Ioan Muresanu and Ziwen Han and Keiran Paster and Silviu Pitis and Harris Chan and Jimmy Ba},
  booktitle = {The Eleventh International Conference on Learning Representations, ICLR},
  year      = {2023}
}

@inproceedings{opro2024,
  title     = {Large Language Models as Optimizers},
  author    = {Chengrun Yang and Xuezhi Wang and Yifeng Lu and Hanxiao Liu and Quoc V. Le and Denny Zhou and Xinyun Chen},
  booktitle = {The Twelfth International Conference on Learning Representations, ICLR},
  year      = {2024}
}

@inproceedings{promptbreeder2023,
  title     = {Promptbreeder: Self-Referential Self-Improvement Via Prompt Evolution},
  author    = {Chrisantha Fernando and Dylan Banarse and Henryk Michalewski and Simon Osindero and Tim Rockt\"aschel},
  booktitle = {Forty-first International Conference on Machine Learning, ICML},
  year      = {2024}
}

@inproceedings{evoprompt2024,
  title     = {EvoPrompt: Connecting LLMs with Evolutionary Algorithms Yields Powerful Prompt Optimizers},
  author    = {Qingyan Guo and Rui Wang and Junliang Guo and Bei Li and Kaitao Song and Xu Tan and Guoqing Liu and Jiang Bian and Yujiu Yang},
  booktitle = {The Twelfth International Conference on Learning Representations, ICLR},
  year      = {2024}
}

@inproceedings{dtvg2025,
  title     = {Dynamic Task Vector Grouping for Efficient Multi-Task Prompt Tuning},
  author    = {Peiyi Zhang and Richong Zhang and Zhijie Nie},
  booktitle = {Findings of the Association for Computational Linguistics: ACL},
  publisher = {Association for Computational Linguistics},
  year      = {2025}
}

@misc{p1prompt2026,
  title         = {p1: Better Prompt Optimization with Fewer Prompts},
  author        = {Zhaolin Gao and Yu Wang and Bo Liu and Thorsten Joachims and Kiant\'{e} Brantley and Wen Sun},
  year          = {2026},
  eprint        = {2604.08801},
  archivePrefix = {arXiv},
  primaryClass  = {cs.LG},
  url           = {https://arxiv.org/abs/2604.08801}
}

@article{voyager2023,
  title   = {Voyager: An Open-Ended Embodied Agent with Large Language Models},
  author  = {Guanzhi Wang and Yuqi Xie and Yunfan Jiang and Ajay Mandlekar and Chaowei Xiao and Yuke Zhu and Linxi Fan and Anima Anandkumar},
  journal = {Transactions on Machine Learning Research},
  year    = {2024}
}

@inproceedings{selfrefine2023,
  title     = {Self-Refine: Iterative Refinement with Self-Feedback},
  author    = {Aman Madaan and Niket Tandon and Prakhar Gupta and Skyler Hallinan and Luyu Gao and Sarah Wiegreffe and Uri Alon and Nouha Dziri and Shrimai Prabhumoye and Yiming Yang and Shashank Gupta and Bodhisattwa Prasad Majumder and Katherine Hermann and Sean Welleck and Amir Yazdanbakhsh and Peter Clark},
  booktitle = {Advances in Neural Information Processing Systems 36, NeurIPS},
  year      = {2023}
}

@inproceedings{reflexion2023,
  title     = {Reflexion: Language Agents with Verbal Reinforcement Learning},
  author    = {Noah Shinn and Federico Cassano and Edward Berman and Ashwin Gopinath and Karthik Narasimhan and Shunyu Yao},
  booktitle = {Advances in Neural Information Processing Systems 36, NeurIPS},
  year      = {2023}
}

@inproceedings{react2023,
  title     = {ReAct: Synergizing Reasoning and Acting in Language Models},
  author    = {Shunyu Yao and Jeffrey Zhao and Dian Yu and Nan Du and Izhak Shafran and Karthik Narasimhan and Yuan Cao},
  booktitle = {The Eleventh International Conference on Learning Representations, ICLR},
  year      = {2023}
}

@inproceedings{sweagent2024,
  title     = {SWE-agent: Agent-Computer Interfaces Enable Automated Software Engineering},
  author    = {John Yang and Carlos E. Jimenez and Alexander Wettig and Kilian Lieret and Shunyu Yao and Karthik Narasimhan and Ofir Press},
  booktitle = {Advances in Neural Information Processing Systems 37, NeurIPS},
  year      = {2024}
}

@inproceedings{instructgpt2022,
  title     = {Training language models to follow instructions with human feedback},
  author    = {Long Ouyang and Jeff Wu and Xu Jiang and Diogo Almeida and Carroll L. Wainwright and Pamela Mishkin and Chong Zhang and Sandhini Agarwal and Katarina Slama and Alex Ray and John Schulman and Jacob Hilton and Fraser Kelton and Luke Miller and Maddie Simens and Amanda Askell and Peter Welinder and Paul Christiano and Jan Leike and Ryan Lowe},
  booktitle = {Advances in Neural Information Processing Systems 35, NeurIPS},
  year      = {2022}
}

@misc{memgpt2023,
  title         = {MemGPT: Towards LLMs as Operating Systems},
  author        = {Charles Packer and Sarah Wooders and Kevin Lin and Vivian Fang and Shishir G. Patil and Ion Stoica and Joseph E. Gonzalez},
  year          = {2023},
  eprint        = {2310.08560},
  archivePrefix = {arXiv},
  primaryClass  = {cs.AI},
  url           = {https://arxiv.org/abs/2310.08560}
}

@inproceedings{dspy2023,
  title     = {DSPy: Compiling Declarative Language Model Calls into State-of-the-Art Pipelines},
  author    = {Omar Khattab and Arnav Singhvi and Paridhi Maheshwari and Zhiyuan Zhang and Keshav Santhanam and Sri Vardhamanan and Saiful Haq and Ashutosh Sharma and Thomas T. Joshi and Hanna Moazam and Heather Miller and Matei Zaharia and Christopher Potts},
  booktitle = {The Twelfth International Conference on Learning Representations, ICLR},
  year      = {2024}
}

@article{funsearch2024,
  title     = {Mathematical discoveries from program search with large language models},
  author    = {Romera-Paredes, Bernardino and Barekatain, Mohammadamin and Novikov, Alexander and Balog, Matej and Kumar, M. Pawan and Dupont, Emilien and Ruiz, Francisco J. R. and Ellenberg, Jordan S. and Wang, Pengming and Fawzi, Omar and Kohli, Pushmeet and Fawzi, Alhussein},
  journal   = {Nature},
  volume    = {625},
  number    = {7995},
  pages     = {468--475},
  year      = {2024},
  publisher = {Nature Publishing Group},
  doi       = {10.1038/s41586-023-06924-6},
  issn      = {1476-4687}
}

@inproceedings{adas2024,
  title     = {Automated Design of Agentic Systems},
  author    = {Shengran Hu and Cong Lu and Jeff Clune},
  booktitle = {The Thirteenth International Conference on Learning Representations, ICLR},
  year      = {2025}
}

@inproceedings{eureka2024,
  title     = {Eureka: Human-Level Reward Design via Coding Large Language Models},
  author    = {Yecheng Jason Ma and William Liang and Guanzhi Wang and De-An Huang and Osbert Bastani and Dinesh Jayaraman and Yuke Zhu and Linxi Fan and Anima Anandkumar},
  booktitle = {The Twelfth International Conference on Learning Representations, ICLR},
  year      = {2024}
}

@misc{alphaevolve2025,
  title         = {{AlphaEvolve}: A coding agent for scientific and algorithmic discovery},
  author        = {Alexander Novikov and Ng{\^a}n V{\~u} and Marvin Eisenberger and Emilien Dupont and Po-Sen Huang and Adam Zsolt Wagner and Sergey Shirobokov and Borislav Kozlovskii and Francisco J. R. Ruiz and Abbas Mehrabian and M. Pawan Kumar and Abigail See and Swarat Chaudhuri and George Holland and Alex Davies and Sebastian Nowozin and Pushmeet Kohli and Matej Balog},
  year          = {2025},
  eprint        = {2506.13131},
  archivePrefix = {arXiv},
  primaryClass  = {cs.AI},
  url           = {https://arxiv.org/abs/2506.13131}
}

@misc{godel2003,
  title         = {Goedel Machines: Self-Referential Universal Problem Solvers Making Provably Optimal Self-Improvements},
  author        = {Juergen Schmidhuber},
  year          = {2003},
  eprint        = {cs/0309048},
  archivePrefix = {arXiv},
  primaryClass  = {cs.LO},
  url           = {https://arxiv.org/abs/cs/0309048},
  note          = {Originally posted 2003; later revised. A book-chapter version appears in ``Artificial General Intelligence,'' Springer, 2007.}
}

@inproceedings{cot2022,
  title     = {Chain-of-Thought Prompting Elicits Reasoning in Large Language Models},
  author    = {Jason Wei and Xuezhi Wang and Dale Schuurmans and Maarten Bosma and Brian Ichter and Fei Xia and Ed Chi and Quoc Le and Denny Zhou},
  booktitle = {Advances in Neural Information Processing Systems 35, NeurIPS},
  year      = {2022}
}

@misc{arcagi2019,
  title         = {On the Measure of Intelligence},
  author        = {Fran\c{c}ois Chollet},
  year          = {2019},
  eprint        = {1911.01547},
  archivePrefix = {arXiv},
  primaryClass  = {cs.AI},
  url           = {https://arxiv.org/abs/1911.01547}
}

@inproceedings{gpqa2023,
  title     = {GPQA: A Graduate-Level Google-Proof Q\&A Benchmark},
  author    = {David Rein and Betty Li Hou and Asa Cooper Stickland and Jackson Petty and Richard Yuanzhe Pang and Julien Dirani and Julian Michael and Samuel R. Bowman},
  booktitle = {First Conference on Language Modeling, COLM},
  year      = {2024}
}

@misc{mbpp2021,
  title         = {Program Synthesis with Large Language Models},
  author        = {Jacob Austin and Augustus Odena and Maxwell Nye and Maarten Bosma and Henryk Michalewski and David Dohan and Ellen Jiang and Carrie Cai and Michael Terry and Quoc Le and Charles Sutton},
  year          = {2021},
  eprint        = {2108.07732},
  archivePrefix = {arXiv},
  primaryClass  = {cs.PL},
  url           = {https://arxiv.org/abs/2108.07732}
}

@inproceedings{mmlu2021,
  title     = {Measuring Massive Multitask Language Understanding},
  author    = {Dan Hendrycks and Collin Burns and Steven Basart and Andy Zou and Mantas Mazeika and Dawn Song and Jacob Steinhardt},
  booktitle = {The Ninth International Conference on Learning Representations, ICLR},
  year      = {2021}
}

@misc{limo2025,
  title         = {LIMO: Less is More for Reasoning},
  author        = {Yixin Ye and Zhen Huang and Yang Xiao and Ethan Chern and Shijie Xia and Pengfei Liu},
  year          = {2025},
  eprint        = {2502.03387},
  archivePrefix = {arXiv},
  primaryClass  = {cs.CL},
  url           = {https://arxiv.org/abs/2502.03387}
}

@misc{d1diff2025,
  title         = {d1: Scaling Reasoning in Diffusion Large Language Models via Reinforcement Learning},
  author        = {Siyan Zhao and Devaansh Gupta and Qinqing Zheng and Aditya Grover},
  year          = {2025},
  eprint        = {2504.12216},
  archivePrefix = {arXiv},
  primaryClass  = {cs.LG},
  url           = {https://arxiv.org/abs/2504.12216}
}

@misc{s1simplescaling2025,
  title         = {s1: Simple Test-Time Scaling},
  author        = {Niklas Muennighoff and Zitong Yang and Weijia Shi and Xiang Lisa Li and Li Fei-Fei and Hannaneh Hajishirzi and Luke Zettlemoyer and Percy Liang and Emmanuel Cand\`es and Tatsunori Hashimoto},
  year          = {2025},
  eprint        = {2501.19393},
  archivePrefix = {arXiv},
  primaryClass  = {cs.CL},
  url           = {https://arxiv.org/abs/2501.19393}
}

@misc{deepseekv32_2025,
  author       = {{DeepSeek-AI}},
  title        = {{DeepSeek-V3.2}: Pushing the Frontier of Open Large Language Models},
  year         = {2025},
  eprint       = {2512.02556},
  archivePrefix = {arXiv},
  primaryClass = {cs.CL},
  url          = {https://arxiv.org/abs/2512.02556}
}

@misc{gemini3pro_2026,
  author       = {{Google DeepMind}},
  title        = {{Gemini 3.1 Pro} Model Card},
  year         = {2026},
  howpublished = {\url{https://storage.googleapis.com/deepmind-media/Model-Cards/Gemini-3-1-Pro-Model-Card.pdf}}
}

@misc{gemini3flashlite_2026,
  author       = {{Google DeepMind}},
  title        = {{Gemini 3.1 Flash-Lite} Model Card},
  year         = {2026},
  howpublished = {\url{https://storage.googleapis.com/deepmind-media/Model-Cards/Gemini-3-1-Flash-Lite-Model-Card.pdf}}
}

@misc{claudeopus46_2026,
  author       = {Anthropic},
  title        = {{Claude Opus 4.6} System Card},
  year         = {2026},
  howpublished = {\url{https://www-cdn.anthropic.com/14e4fb01875d2a69f646fa5e574dea2b1c0ff7b5.pdf}}
}

@misc{aime2025hf,
  author       = {{OpenCompass Contributors}},
  title        = {{AIME} 2025},
  year         = {2025},
  howpublished = {\url{https://huggingface.co/datasets/opencompass/AIME2025}}
}
